\documentclass[10pt,journal,compsoc]{IEEEtran}

\usepackage{color}
\usepackage{epsfig}
\usepackage{graphicx}
\usepackage{amsmath}
\usepackage{amssymb}
\usepackage{multirow}
\usepackage{epstopdf}
\usepackage{subfigure}
\usepackage{url}
\usepackage{booktabs}
\usepackage{algorithm}
\usepackage{algpseudocode}

\newcommand{\etal}{\textit{et al.}}

\newcommand{\eg}{\textit{e.g.}}
\newcommand{\ie}{\textit{i.e.}}
\ifCLASSOPTIONcompsoc
  \usepackage[nocompress]{cite}
\else
  \usepackage{cite}
\fi

\begin{document}

\title{GP-UNIT: Generative Prior for Versatile Unsupervised Image-to-Image Translation}

\author{Shuai~Yang,~\IEEEmembership{Member,~IEEE},
        Liming Jiang,~\IEEEmembership{Student Member,~IEEE},\\
        Ziwei Liu,~\IEEEmembership{Member,~IEEE},
        and~Chen Change Loy,~\IEEEmembership{Senior Member,~IEEE}
\thanks{
This work is supported under the RIE2020 Industry Alignment Fund - Industry Collaboration Projects (IAF-ICP) Funding Initiative, as well as cash and in-kind
contribution from the industry partner(s). It is also supported by MOE AcRF Tier 2 (T2EP20221-0011, T2EP20221-0012) and NTU NAP grant.}
\thanks{Corresponding author: Chen Change Loy.}
\thanks{Shuai Yang, Liming Jiang, Ziwei Liu and Chen Change Loy are with S-Lab,
Nanyang Technological University (NTU), Singapore 639798.
(E-mail: \{shuai.yang, liming002, ziwei.liu, ccloy\}@ntu.edu.sg).}
}
\markboth{IEEE TRANSACTIONS ON PATTERN ANALYSIS AND MACHINE INTELLIGENCE}%
{Shell \MakeLowercase{\textit{et al.}}: Bare Advanced Demo of IEEEtran.cls for IEEE Computer Society Journals}

\IEEEtitleabstractindextext{
\begin{abstract}
  Recent advances in deep learning have witnessed many successful unsupervised image-to-image translation models that learn correspondences between two visual domains without paired data. However, it is still a great challenge to build robust mappings between various domains especially for those with drastic visual discrepancies. In this paper, we introduce a novel versatile framework, Generative Prior-guided UNsupervised Image-to-image Translation (\textbf{GP-UNIT}), that improves the quality, applicability and controllability of the existing translation models. The key idea of GP-UNIT is to distill the generative prior from pre-trained class-conditional GANs to build coarse-level cross-domain correspondences, and to apply the learned prior to adversarial translations to excavate fine-level correspondences. With the learned multi-level content correspondences, GP-UNIT is able to perform valid translations between both close domains and distant domains. For close domains, GP-UNIT can be conditioned on a parameter to determine the intensity of the content correspondences during translation, allowing users to balance between content and style consistency. For distant domains, semi-supervised learning is explored to guide GP-UNIT to discover accurate semantic correspondences that are hard to learn solely from the appearance.  We validate the superiority of GP-UNIT over state-of-the-art translation models in robust, high-quality and diversified translations between various domains through extensive experiments.
\end{abstract}

\begin{IEEEkeywords}
Multi-level correspondence, prior distillation, coarse-to-fine, distant domains, multi-modal translation.
\end{IEEEkeywords}}

\maketitle

\IEEEdisplaynontitleabstractindextext
\IEEEpeerreviewmaketitle

\ifCLASSOPTIONcompsoc
\IEEEraisesectionheading{\section{Introduction}\label{sec:introduction}}
\else
\section{Introduction}
\fi

\IEEEPARstart{T}{HE} goal of unsupervised image-to-image translation (UNIT) is to map images between two visual domains without paired data.
In the mainstream UNIT framework, cross-domain mappings are mostly built via cycle-consistency~\cite{Zhu2017Unpaired}, which assumes a bijection between two domains. Although this framework has achieved good results on simple cases like female-to-male translation, such assumption is often too restrictive for real-world diverse domains. The performance often degrades severely in translations with drastic shape and appearance discrepancies, such as translating human faces to animal faces, limiting the practical applications.

Translating across various domains requires one to establish the translation at multiple and adaptive semantic levels.
Thanks to our cognitive abilities, humans can easily accomplish this task.
For instance, to translate a human face to a cat face, one can use the more reliable correspondence of facial components such as the eyes between a human and a cat rather than on the local textures.
In the more extreme case of distant domains, such as animals and man-made objects, a translation is still possible if their correspondence can be determined at a higher abstract semantic level, for example through affirming the frontal orientation of an object or the layout of an object within the image.
Therefore, we are interested in the following research question: \textit{Can a machine learn to discover appropriate correspondences in adaptive semantic levels to different domains just like humans, so as to build a versatile and robust UNIT framework for various tasks?}

\begin{figure*}[t]
\centering
\includegraphics[width=0.97\linewidth]{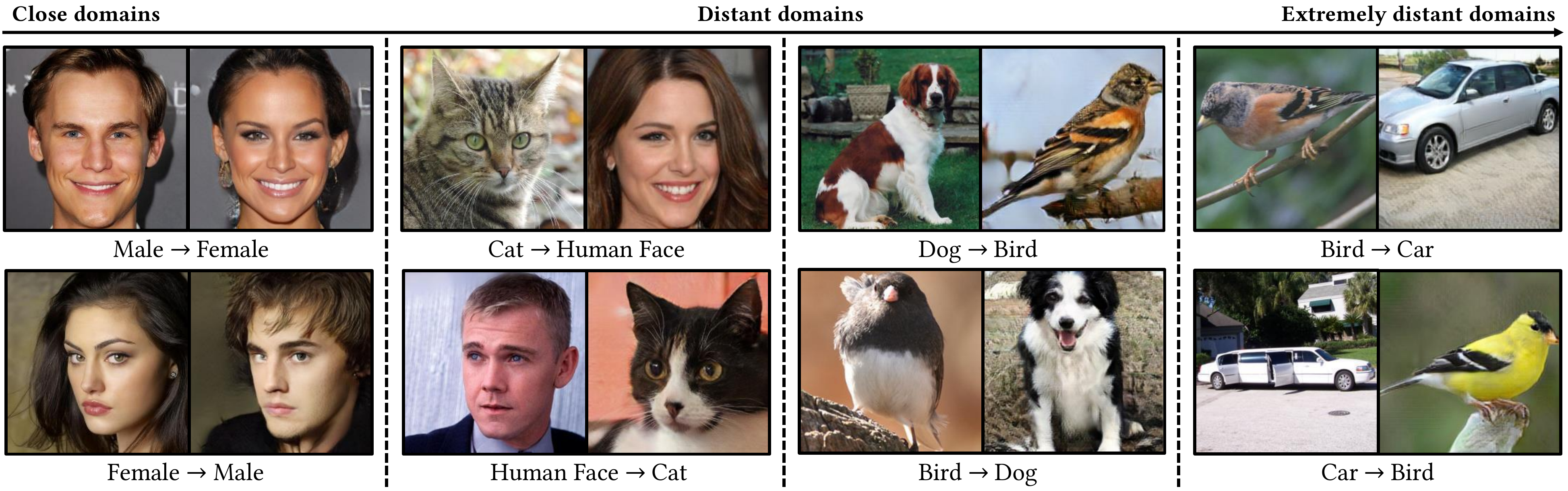}\vspace{-2mm}
\caption{We propose a versatile unsupervised image translation framework with generative prior that supports various translations from close domains to distant domains with drastic shape and appearance discrepancies. Each group shows \textit{(left)} the input and \textit{(right)} our result.}\vspace{-1mm}
\label{fig:teaser}
\end{figure*}

\begin{figure}[t]
\centering
\includegraphics[width=\linewidth]{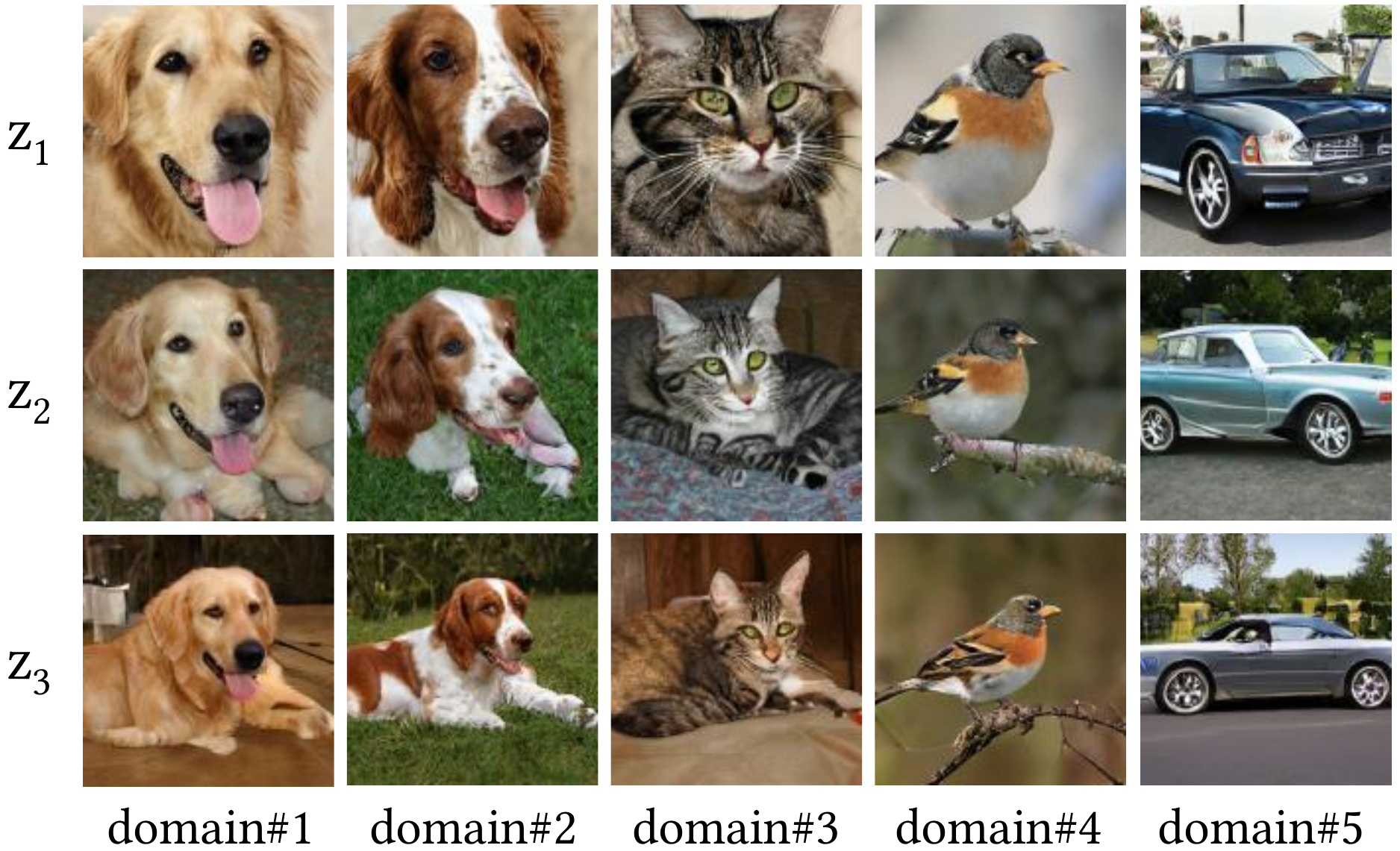}\vspace{-2mm}
\caption{Generative space of BigGAN~\cite{brock2018large}. Objects of different classes generated from the same latent code have a high degree of content correspondences.}\vspace{-1mm}
\label{fig:analysis}
\end{figure}

The abilities of humans in making associations across diverse domains are largely gained through experiences, namely with priors, \eg, seeing many relevant instances in the past. We believe machines need priors too, to accomplish such tasks.
In this work, we introduce a versatile and robust UNIT model through a novel use of generative prior and achieve promising results as shown in Fig.~\ref{fig:teaser}. Specifically, we show that a class-conditional GAN, such as BigGAN~\cite{brock2018large}, provides powerful hints on how different objects are linked -- objects of different classes generated from the same latent code have a high degree of content correspondences (Fig.~\ref{fig:analysis}).
Through generating pairs of such cross-domain images, we can mine the unique prior of the class-conditional GAN and use them to guide an image translation model in building effective and adaptable content mappings across various classes (we will use ``domain'' instead of ``class'' hereafter).

However, such prior is not immediately beneficial to UNIT.
BigGAN, by nature, covers a large number of domains, which makes it an ideal choice of prior for our problem to achieve translation between multiple domains. However, the coverage of many domains inevitably limits the quality and intra-domain diversity of the captured distribution of each domain. Without a careful treatment, such a limitation will severely affect the performance of UNIT in generating high-quality and diverse results.

To overcome the problem above, we decompose a translation task into coarse-to-fine stages: 1) generative prior distillation to learn robust cross-domain correspondences at a high semantic level and 2) adversarial image translation to build finer adaptable correspondences at multiple semantic levels. In the first stage, we train a content encoder to extract disentangled content representation by distilling the prior from the content-correlated data generated by BigGAN.
In the second stage, we apply the pre-trained content encoder to the specific translation task, independent of the generative space of BigGAN,
and propose a dynamic skip connection module to learn adaptable correspondences,
so as to yield plausible and diverse translation results.

To our knowledge, this is the first work to employ BigGAN generative prior for unsupervised\footnote{Following the definition in Liu \etal~\cite{Liu2017Unsupervised}, we call our method unsupervised since our method and the pre-trained BigGAN only use the marginal distributions in individual domains without any explicit cross-domain correspondence supervision.} image-to-image translation.
In particular, we propose a versatile Generative Prior-guided UNsupervised Image-to-image Translation framework (\textbf{GP-UNIT}) to expand the application scenarios of previous UNIT methods that mainly handles close domains.~Our framework shows positive improvements over previous cycle-consistency-guided frameworks in:
1) capturing coarse-level correspondences across various heterogeneous and asymmetric domains, beyond the ability of cycle-consistency guidance;
2) learning fine-level correspondences applicable to various tasks adaptively; and
3) retaining essential content features in the coarse-to-fine stages, avoiding artifacts from the source domain commonly observed in cycle reconstruction.

Compared with our previous work~\cite{yang2022unsupervised},
we further explore the degree control of content consistency and semi-supervised learning to strengthen the specialization of GP-UNIT for specific tasks.
For translation tasks between close domains, we introduce a controllable version of GP-UNIT with a new parameter $\ell$ to adjust how many content features are preserved, which allows users to flexibly balance between the content and style.
For translation tasks between distant domains, we propose an economic semi-supervised learning to train GP-UNIT to discover challenging cross-domain correspondences that are hard to learn solely from objects' appearance.
In addition, comprehensive experiments are conducted to analyze the image translation performance of the extended GP-UNIT,
including qualitative and quantitative evaluations on the effects of $\ell$ in content preservation, and the comparison results with and without the proposed semi-supervised learning.
In summary, our contributions are fivefold:
\begin{itemize}
  \item We propose a versatile GP-UNIT framework that promotes the overall quality and applicability of UNIT with BigGAN generative prior.
  \item We present an effective way of learning robust correspondences across non-trivially distant domains at a high semantic level via generative prior distillation.
  \item We design a novel coarse-to-fine scheme to learn cross-domain correspondences adaptively at different semantic levels.
  \item We propose a content consistency controllable framework to achieve a trade-off between the content and style during image translation.
  \item We present a semi-supervised learning strategy to improve the accuracy of the cross-domain correspondences by combining generative prior and discriminative prior.
\end{itemize}

\begin{figure*}[t]
\centering
\includegraphics[width=0.93\linewidth]{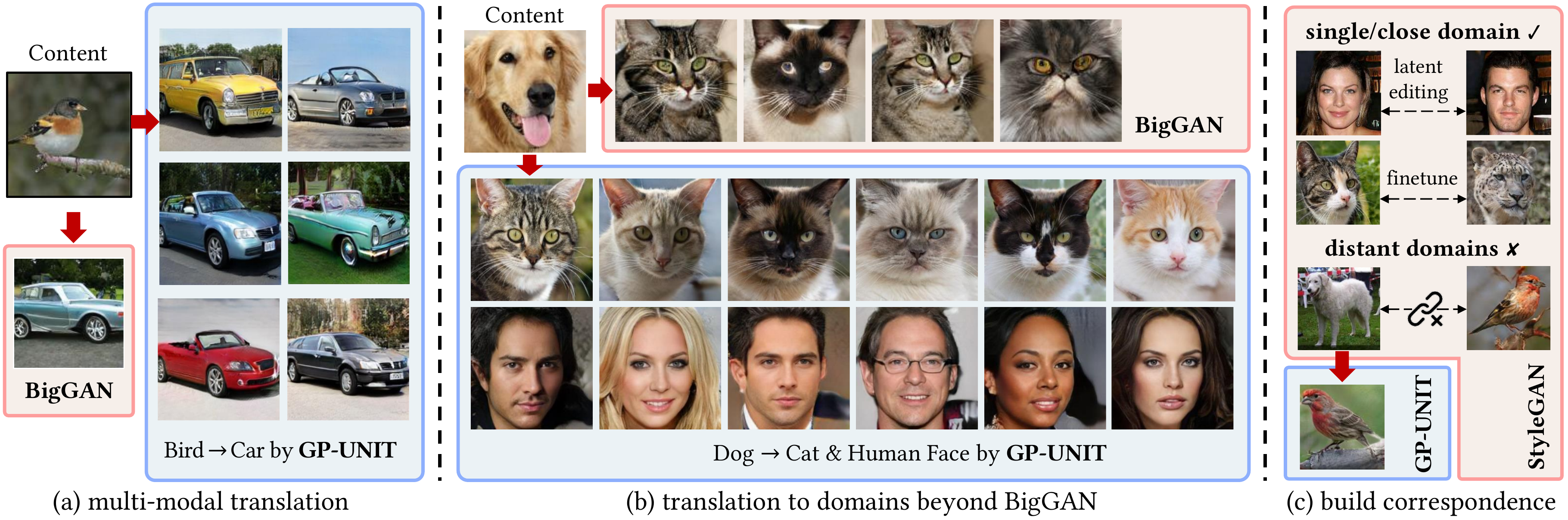}\vspace{-2mm}
\caption{Comparison of the generative spaces of BigGAN, StyleGAN and GP-UNIT. GP-UNIT realizes (a) multi-modal translation, (b) generates cats and human faces beyond ImageNet, and (c) builds robust mappings between distant domains. StyleGAN images are from \protect\cite{shen2020interpreting,kwong2021unsupervised}. Our results in (a)(b) are generated based on the content image and randomly sampled style features.}\vspace{-1mm}
\label{fig:comparison_biggan}
\end{figure*}

The rest of this paper is organized as follows.
In Section~\ref{sec:related_work}, we review related studies in unsupervised image-to-image translation and adversarial image generation.
Section~\ref{sec:analysis} analyzes the cross-domain correspondences prior of BigGAN and presents a content encoder to distill the prior.
In Section~\ref{sec:translation}, the details of the proposed generative prior-guided translation model are introduced.
Section~\ref{sec:experiment} validates the superiority of our method via extensive experiments and comparisons with state-of-the-art UNIT methods.
Finally, we conclude our work in Section~\ref{sec:conclusion}.

\section{Related Work}
\label{sec:related_work}
\vspace{-1mm}

\noindent
\textbf{Unsupervised image-to-image translation.}
To learn the mapping between two domains without supervision, CycleGAN~\cite{Zhu2017Unpaired} proposes a novel cycle consistency constraint to build a bi-directional relationship between domains.
To better capture domain-invariant features, representation disentanglement has been investigated extensively in UNIT,
where a content encoder and a style encoder~\cite{Liu2017Unsupervised,Choi2017StarGAN,huang2018multimodal,choi2020stargan,liu2019few,jiang2020tsit} are usually employed to extract domain-invariant content features and domain-specific style features, respectively.
However, learning a disentangled representation between two domains with drastic differences is non-trivial.
To cope with the large visual discrepancy, COCO-FUNIT~\cite{saito2020coco} designs a content-conditioned style encoder to prevent the translation of task-irrelevant appearance information. TGaGa~\cite{wu2019transgaga} uses landmarks to build geometry mappings.
StyleD~\cite{kim2022style} combines the style encoder and the discriminator to learn a precise style space to improve disentanglement.
TraVeLGAN~\cite{amodio2019travelgan} proposes a siamese network to seek shared semantic features across domains, and
U-GAT-IT~\cite{kim2019u} leverages an attention module to focus on important regions distinguishing between two domains.
These methods struggle to seek powerful and balanced domain-related representation for specific domains so are less adaptive to the various translation tasks, inevitably failing in certain cases.
CUT~\cite{park2020contrastive} and Hneg-SRC~\cite{jung2022exploring} leverage patch-level contrastive learning to learn domain correspondence, which is less effective for large-scale shape transformation.
Ojha~\etal~\cite{ojha2021generating} proposes to decompose images into four factors of background, pose, shape and appearance (color and texture), and learn to recombine cross-domain shape and appearance with contrastive learning. The disentanglement enables it to handle distant domains like birds and cars, but is limited to appearance transfer such as rendering a car with a bird's color. It cannot transform the object shape from one domain to another like our method.
Different from these methods, we propose a new coarse-to-fine scheme -- coarse-level cross-domain content correspondences at a highly abstract semantic level are first built, based on which fine-level correspondences adaptive to the task are gradually learned. Such a scheme empowers us to build robust mappings to handle various tasks.

\noindent
\textbf{Adversarial image generation.}
Generative Adversarial Network (GAN)~\cite{goodfellow2014generative} introduces a discriminator to compete with the generator to adversarially approximate the real image distribution.
Among various models, StyleGAN~\cite{karras2019style,karras2020analyzing} has shown promising results.
Many works~\cite{collins2020editing,zhu2020domain,chan2020glean,patashnik2021styleclip,jiang2021talk} exploit the generative prior from StyleGAN to ensure superior image quality by restricting the modulated image to be within the generative space of StyleGAN. However, StyleGAN is an unconditional GAN that is limited to a single domain or close domains~\cite{kwong2021unsupervised}.
BigGAN~\cite{brock2018large} is able to synthesize images in different domains but at the expense of quality and intra-domain diversity. Thus it is not straightforward to exploit BigGAN prior following the aforementioned works.
To circumvent this limitation, in this paper, we distill the generative prior from content-correlated data generated by BigGAN and apply it to the image translation task to generate high-quality images.

\section{Generative Prior Distillation}
\label{sec:analysis}
\vspace{-1mm}

\subsection{Cross-Domain Correspondences Prior}

Our framework is motivated by the following observation~\cite{AlyafeaiGradient2018Gans,harkonen2020ganspace} -- objects generated by BigGAN, despite originating from different domains, share high content correspondences when generated from the same latent code.
Figure~\ref{fig:analysis} shows the generative space of BigGAN characterized by three latent codes ($z_1$, $z_2$, $z_3$) across five domains.
For each latent code, fine-grained correspondences can be observed between semantically related dogs and cats, such as the face features and body postures.
For birds and vehicles, which are rather different, one can also observe coarse-level correspondences in terms of orientation and layout.

The interesting phenomenon suggests that \textit{there is an inherent content correspondence at a highly abstract semantic level regardless of the domain discrepancy in the BigGAN generative space}.
In particular, objects with the same latent code share either the same or very similar abstract representations in the first few layers, based on which domain-specific details are gradually added.

\begin{figure*}[t]
\centering
\includegraphics[width=0.93\linewidth]{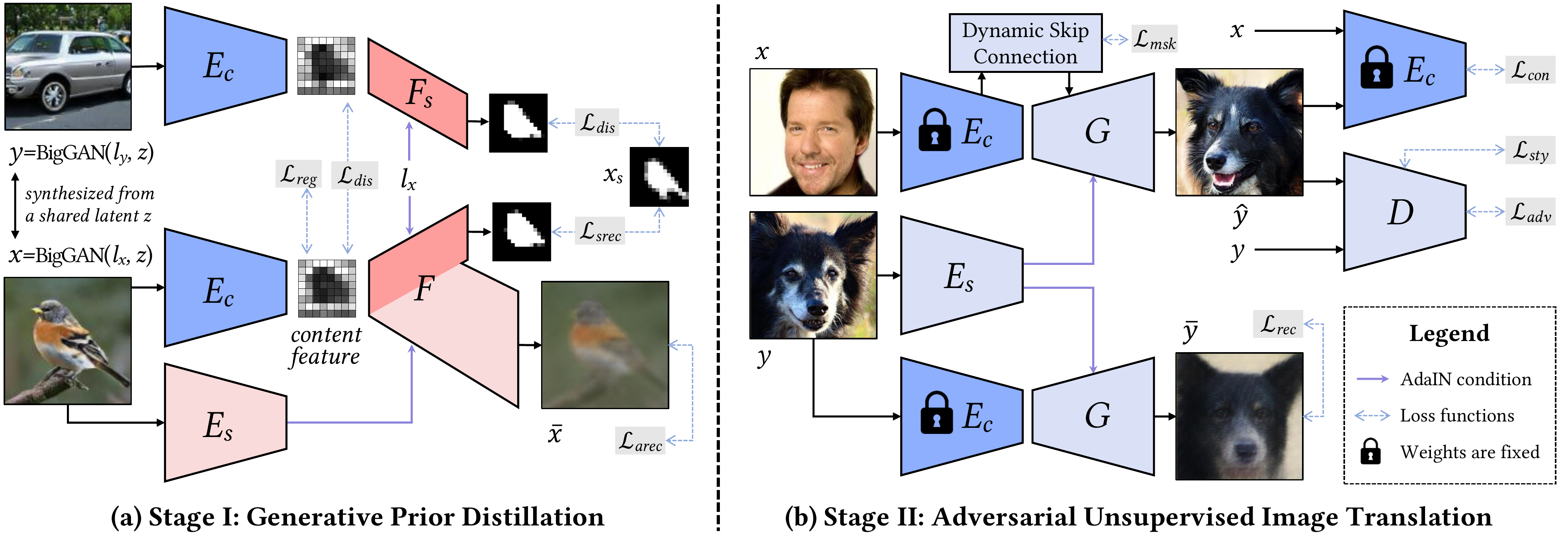}\vspace{-2mm}
\caption{Overview of the proposed GP-UNIT. In the first stage, we use a content encoder $E_c$ to extract shared coarse-level content features between a pair correlated images $(x,y)$ generated by BigGAN from a common random latent code in two random domains. In the second stage, we build our translation network based on the content encoder $E_c$ in the first stage. For simplicity, we omit the classifier $C$.}\vspace{-1mm}
\label{fig:framework}
\end{figure*}

In this paper, we exploit this generative prior for building robust mappings and
choose BigGAN for its rich cross-domain prior. Nevertheless, its generative space is limited in \textbf{quality} and \textbf{diversity} for our purpose.
In terms of quality, BigGAN sometimes generates unrealistic objects, such as the dog body with $z_2$ in Fig.~\ref{fig:analysis}.
As for diversity, first, the space lacks intra-domain variation, \eg, the diversity in textures of a dog or colors of a bird in the same domain are pretty limited. Using such prior in UNIT will overfit the model to limited appearances.
Second, the BigGAN generative space is limited to 1,000 domains of ImageNet~\cite{russakovsky2015imagenet}, which is inadequate for an actual UNIT. For instance, it only has four kinds of domestic cats as in Fig.~\ref{fig:comparison_biggan}(b) and excludes the important domain of human faces.

StyleGAN is not suitable for our task despite its generative space is of high quality and diversity. This is because it is limited to a single domain, and thus it is mainly applied to attribute transfer within one domain via latent editing~\cite{collins2020editing,zhu2020domain,patashnik2021styleclip}.
Recently, cross-domain translations on StyleGAN have been achieved via finetuning~\cite{pinkney2020resolution,kwong2021unsupervised}, but this still assumes a small distance between the models of the source and target domains and is therefore still restricted to close domains.
The assumption makes StyleGAN prior less applicable to more complex translation tasks.

Our framework solves the above problems by distilling a general generative prior from BigGAN instead of directly constraining the latent or image space. It enables us to design and train the translation module independently. Therefore, we can realize multi-modal translation (Fig.~\ref{fig:comparison_biggan}(a)), generalize to classes beyond ImageNet (Fig.~\ref{fig:comparison_biggan}(b)) and build robust mappings between distant domains (Fig.~\ref{fig:comparison_biggan}(c)). Next, we detail how we distill the prior.

\subsection{Prior Distillation with Content Encoder}

Given the correlated images $(x,y)$ generated by BigGAN from a common random latent code in two random domains $\mathcal{X}$ and $\mathcal{Y}$, our main goal is to train a content encoder $E_c$ to extract their shared coarse-level content features, which can be used to reconstruct their shape and appearance.
Figure~\ref{fig:framework}(a) illustrates this autoencoder pipeline for generative prior distillation.

Specifically, we use a decoder $F$ to recover the appearance $x$ based on its content feature $E_c(x)$, style feature $E_s(x)$ extracted by a style encoder $E_s$ and the domain label $l_x$.
We further exploit the shallow layers $F_s$ of $F$ to predict the shape of $x$ (\textit{i.e.},  instance segmentation map $x_s$, which is extracted from $x$ by HTC~\cite{chen2019hybrid}) based on $E_c(x)$ and $l_x$. We find such auxiliary prediction eases the training on hundreds of domains. Besides the shape and appearance reconstruction, we further regularize the content feature in three ways for disentanglement: 1) $x$ and $y$ should share the same content feature; 2) We introduce a classifier $C$ with a gradient reversal layer $R$~\cite{ganin2015unsupervised} to make the content feature domain-agnostic; 3) We limit $E_c(x)$ to one channel to eliminate domain information~\cite{sushko2021one} and add Gaussian noise of a fixed variance for robustness. Our objective function is:
\begin{equation}\label{eq:total_loss2}
  \min_{E_c,E_s,F,C}\mathcal{L}_{arec}+\mathcal{L}_{srec}+\mathcal{L}_{dis}+\mathcal{L}_{reg},
\end{equation}
where $\mathcal{L}_{arec}$ is the appearance reconstruction loss measuring the $L_2$ and perceptual loss~\cite{Johnson2016Perceptual} between $\bar{x}=F(E_c(x),E_s(x),l_x)$ and $x$. The shape reconstruction loss, $\mathcal{L}_{srec}$, is defined as
\begin{equation}
  \mathcal{L}_{srec}=\lambda_{s}\mathbb{E}_{x}[\|F_s(E_c(x),l_x)-x_s\|_1].
\end{equation}
The binary loss $\mathcal{L}_{dis}$ with paired inputs narrows the distance between the content features of $x$ and $y$. In addition, we would like to recover the shape of $x$ with the content feature of $y$, which simulates translations:
\begin{equation}
  \mathcal{L}_{dis}=\mathbb{E}_{(x,y)}[\|E_c(x)-E_c(y)\|_1+\lambda_{s}\|F_s(E_c(y),l_x)-x_s\|_1].  \nonumber
\end{equation}
Finally, $\mathcal{L}_{reg}$ guides $C$ to maximize the classification accuracy and pushes $E_c$ to confuse $C$, so that the content feature is domain-agnostic following~\cite{ganin2015unsupervised}.
An $L_2$ norm is further applied to the content feature:
\begin{equation}
  \mathcal{L}_{reg}=\mathbb{E}_{x}[-l_x\log C(R(E_c(x)))]+\lambda_{r}\mathbb{E}_{x}[\|E_c(x)\|_2].  \nonumber
\end{equation}
For unary losses of $\mathcal{L}_{arec}$, $\mathcal{L}_{srec}$ and $\mathcal{L}_{reg}$, we also use real images of ImageNet~\cite{russakovsky2015imagenet} and CelebA-HQ~\cite{karras2018progressive} for training to make $E_c$ more generalizable.

\section{Adversarial Image Translation}
\label{sec:translation}

Given a fixed content encoder $E_c$ pre-trained in the first stage, we build our translation network following a standard style transfer paradigm in the second stage. Thanks to the pre-trained $E_c$ that provides a good measurement for content similarity, our framework does not need cycle training.

As shown in Fig.~\ref{fig:framework}(b), our translation network receives a content input $x\in\mathcal{X}$ and a style input $y\in\mathcal{Y}$. The network extracts their content feature $E_c(x)$ and style feature $E_s(y)$, respectively. Then, a generator $G$ modulates $E_c(x)$  to match the style of $y$ via AdaIN~\cite{huang2017adain}, and finally produces the translated result $\hat{y}=G(E_c(x),E_s(y))$. The realism of $\hat{y}$ is reinforced through an adversarial loss~\cite{goodfellow2014generative}  with a discriminator $D$,
\begin{equation}
  \mathcal{L}_{adv}=\mathbb{E}_{y}[\log D(y)]+\mathbb{E}_{x,y}[\log (1-D(\hat{y}))].
\end{equation}
In addition, $\hat{y}$ is required to fit the style of $y$, while preserving the original content feature of $x$, which can be formulated as a style loss $\mathcal{L}_{sty}$ and a content loss $\mathcal{L}_{con}$,
\begin{align}\label{eq:content_loss}
  \mathcal{L}_{sty}&=\mathbb{E}_{x,y}[\|f_D(\hat{y})-f_D(y)\|_1],\\\label{eq:style_loss}
  \mathcal{L}_{con}&=\mathbb{E}_{x,y}[\|E_c(\hat{y})-E_c(x)\|_1],
\end{align}
where $f_D$ is the style feature defined as the channel-wise mean of the middle layer feature of $D$ following the style definition in~\cite{huang2017adain}.

\subsection{Dynamic Skip Connection}

Domains that are close semantically would usually exhibit fine-level content correspondences that cannot be characterized solely by the abstract content feature. To solve this problem, we propose a dynamic skip connection module, which passes middle layer features $f_E$ from $E_c$ to $G$ and predicts masks $m$ to select the valid elements for establishing fine-level content correspondences.

Our dynamic skip connection is inspired by the GRU-like selective transfer unit~\cite{liu2019stgan}.
Let the superscript $l$ denote the layer of $G$.
The mask $m^l$ at layer $l$ is determined by the encoder feature $f_E^l$ passed to the same layer and a hidden state $h^{l-1}$ at the last layer.
Specifically, we first set the first hidden state $h^0=E_c(x)$ and use the upsampling convolution to match the dimension of $h^{l-1}$ with $f_E^l$ as $\hat{h}^{l-1}=\sigma(W^l_h\circ \uparrow h^{l-1})$, where $\uparrow$, $\circ$ and $W^l_h$ are the upsample operator, convolution operator and convolution weights, respectively. The activation layer is denoted as $\sigma$. Then, our module at layer $l$ updates the hidden state $h^l$ and the encoder feature $\hat{f}_E^l$, and fuses $\hat{f}_E^l$ with the generator feature $f_G^l$ with the predicted mask $m^l$:
\begin{align}
  r^l=\sigma(W^l_r\circ [\hat{h}^{l-1},f_E^l]),&~~~
  m^l=\sigma(W^l_m\circ [\hat{h}^{l-1},f_E^l]),   \nonumber\\
  h^l=r^l\hat{h}^{l-1},&~~~
  \hat{f}_E^l=\sigma(W^l_E\circ [h^l,f_E^l]),   \nonumber\\
  f^l=(1-&m^l)f_G^l+m^l\hat{f}_E^l,   \nonumber
\end{align}
where $[\cdot,\cdot]$ denotes concatenation. $m^l$ has the same dimension of $f_G^l$, serving both channel attention and spatial attention.
Moreover, we apply $L_1$ norm to $m^l$ to make it sparse,
\begin{equation}
  \mathcal{L}_{msk}=\mathbb{E}_{x}\Big[\sum\nolimits_l\|m^l\|_1\Big],
\end{equation}
so that only the most useful content cues from the source domain are selected.

\noindent
\textbf{Full objectives.}
Combining the aforementioned losses, our full objectives take the following form:
\begin{equation}\label{eq:total_loss}
  \min_{G,E_s}\max_{D}\mathcal{L}_{adv}+\lambda_{1}\mathcal{L}_{con}+\lambda_{2}\mathcal{L}_{sty}
  +\lambda_{3}\mathcal{L}_{msk}+\lambda_{4}\mathcal{L}_{rec}. \nonumber
\end{equation}
A reconstruction loss $\mathcal{L}_{rec}$ is added to measure the $L_1$ and perceptual loss~\cite{Johnson2016Perceptual} between $y$ and $\bar{y}=G(E_c(y),E_s(y))$.
Intuitively, we would like the learned style feature of an image to precisely reconstruct itself with the help from its content feature, which stabilizes the network training.

\noindent
\textbf{Style sampling.}
To sample latent style features directly for multi-modal generation without the style images, we follow the post-processing of~\cite{meshry2021step} to train a mapping network to map the unit Gaussian noise to the latent style distribution using a maximum likelihood
criterion~\cite{hoshen2019non}. Please refer to~\cite{hoshen2019non} for the details.

\begin{figure}[t]
\centering
    \includegraphics[width=1\linewidth]{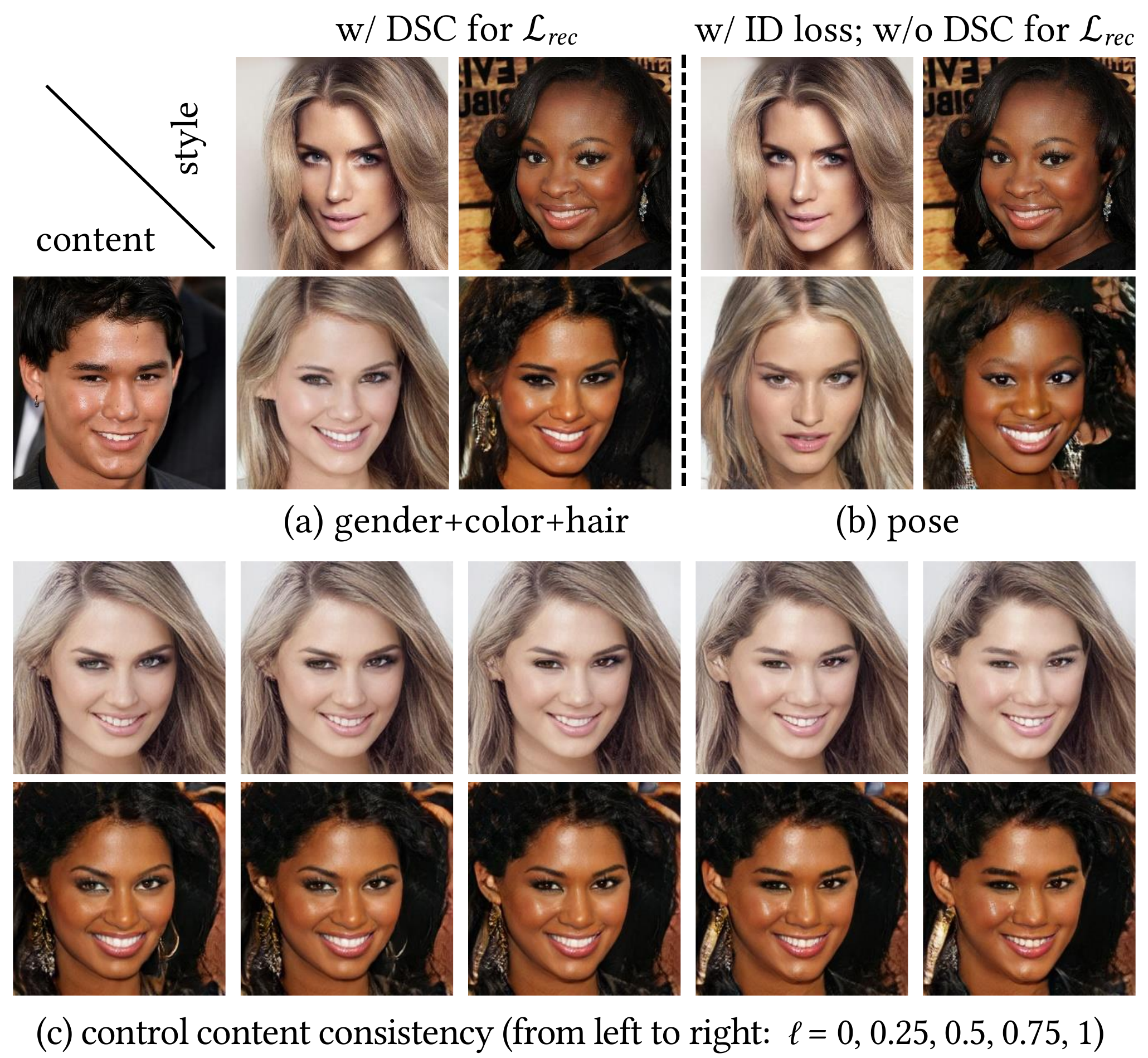}\vspace{-2mm}
\caption{Control the degree of content consistency. (a)(b) The loss function and dynamic skip connection (DSC) are two factors affecting the content consistency. GP-UNIT can be trained under two settings to realize two degrees of content consistency.  (c) We propose a controllable version of GP-UNIT that can flexibly adjust the degree of content consistency with a parameter $\ell$ within a single model.}\vspace{-1mm}
\label{fig:multilevel}
\end{figure}

\subsection{Control Content Consistency for Close Domains}

In UNIT, the degree of content consistency between the content image and the translated image is mainly determined by the definition of the content (features to be preserved) and the style (features to be altered). However, both concepts are ambiguous for different tasks and users, especially for close domains like Male $\leftrightarrow$ Female, \eg, whether the human identity is considered as the style to be transferred. It is hard to give a universal definition that satisfies all users. In this section, we propose a controllable framework that provides a parameter $\ell\in[0,1]$ to control the degree of content consistency during translation. A small $\ell$ is inclined to transfer more style, and a large $\ell$ tends to retain more content features, as illustrated in Fig.~\ref{fig:multilevel}(c). Users can navigate around the generative space under different $\ell$ to select the ideal result that best matches their definition of the content and the style.

In GP-UNIT, there are two factors affecting the content consistency: 1) loss function and 2) dynamic skip connection (DSC).
For example, as shown in Fig.~\ref{fig:multilevel}(a), using DSC to compute $\bar{y}$ in $\mathcal{L}_{rec}$ helps preserve the identity of the content face, which is suitable for gender and color transfer. Meanwhile, in Fig.~\ref{fig:multilevel}(b), when we add an identity loss~\cite{deng2019arcface} between the translated image and the style image, and do not use DSC for $\mathcal{L}_{rec}$ (by setting $m^l$ to an all-zero tensor), most attributes of the style face except pose are transferred, which is suitable for pose transfer.
To introduce such adjustment of content consistency into a single GP-UNIT model, our key idea is to condition DSC on $\ell$ to pass the adaptive content features according to the degree of content consistency.

Specifically, we maintain two groups of parameters in DSC for predicting masks and updating encoder features: $W^l_{m0}$, $W^l_{m1}$, $W^l_{E0}$, $W^l_{E1}$, $a^l_{tran}$, $a^l_{rec}$. The subscripts $0$ and $1$ indicate the extreme case of $\ell=0$ and $\ell=1$, respectively. The new $a^l_{task} (task\in\{tran,rec\})$ are learnable channel-wise attention vectors for computing $\hat{y}$ and $\bar{y}$ during training. The mask $m^l$ and the updated encoder feature $\hat{f}_E^l$ conditioned on $\ell$ are obtained as:
\begin{equation}
\begin{aligned}
  m^l=a^l_{task}&\Big((1-\ell)\sigma(W^l_{m0}\circ [\hat{h}^{l-1},f_E^l])\\
  &+\ell\sigma(W^l_{m1}\circ [\hat{h}^{l-1},f_E^l])\Big).
\end{aligned}
\end{equation}
\begin{equation}
  \hat{f}_E^l=(1-\ell)\sigma(W^l_{E0}\circ [h^l,f_E^l])+\ell\sigma(W^l_{E1}\circ [h^l,f_E^l]).
\end{equation}
In the original GP-UNIT, we use $m^l$ in a binary fashion, \ie, using it as is or setting it to an all-zero tensor. In our controllable version, $a^l_{task}$ is learned to make $m^l$ adaptive to the translation task and reconstruction task.

The conditional DSC is driven by a new content consistency loss $\mathcal{L}_{cc}$. $\mathcal{L}_{cc}$ is similar to the content loss but uses the middle-layer encoder features $f_E^l$ and is weighted by $\ell$
\begin{equation}\label{eq:cc1}
  \mathcal{L}_{cc}=\mathbb{E}_{x,y}\Big[\ell\sum\nolimits_l\lambda_{cc}^l\|f_E^l(\hat{y})-f_E^l(x)\|^2_2\Big],
\end{equation}
where $\lambda_{cc}^l$ is the weight at layer $l$.
For Male $\leftrightarrow$ Female task, we use the identity loss~\cite{deng2019arcface} as a more robust metric
\begin{equation}\label{eq:cc2}
  \mathcal{L}_{cc}=\mathbb{E}_{x,y}[\lambda_{cc}\ell~\text{ID}(\hat{y},x)+(1-\ell)\text{ID}(\hat{y},y)],
\end{equation}
where $\text{ID}(\cdot,\cdot)$ is the identity loss.

\subsection{Semi-Supervised Translation for Distant Domains}
\label{sec:semi}

\begin{figure}[t]
\centering
\includegraphics[width=1\linewidth]{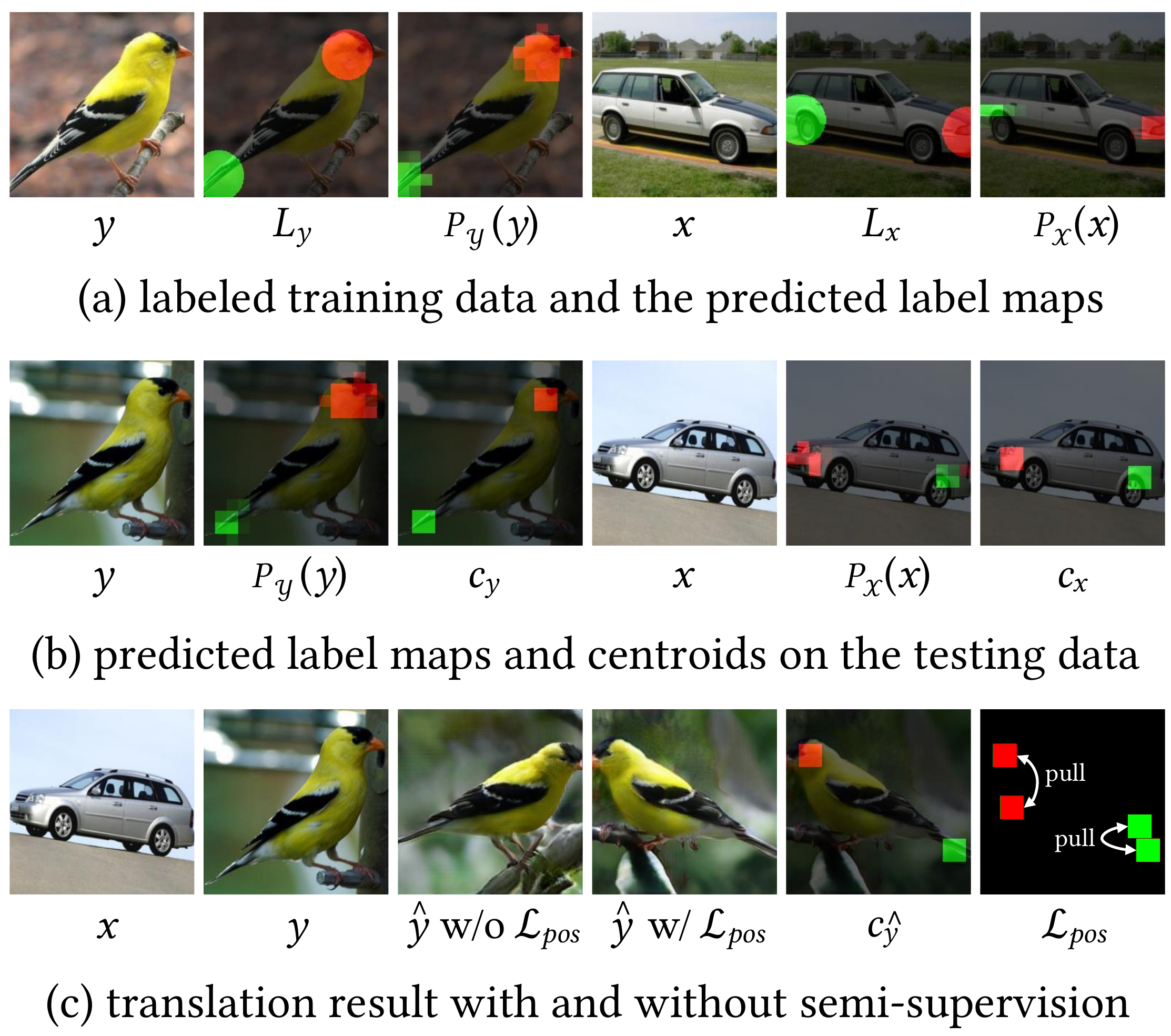}\vspace{-2mm}
\caption{Illustration of semi-supervised translation. (a) We label a small amount of training data to train networks to predict the label map. The trained networks provide supervision on the correspondence beyond appearance. (b) The robust centroid is calculated from the predicted less precise label map. (c) The proposed $\mathcal{L}_{pos}$ penalizes the centroid changes during translation, guiding GP-UNIT to find correct correspondences between the bird head and car front. We overlay the label maps and centroids on the input image for better visualization.}
\vspace{-1mm}
\label{fig:semi}
\end{figure}

It is hard to learn certain semantic correspondences between distant domains solely from their appearance, such as which side of a car is its front.
As shown in Fig.~\ref{fig:semi}(c), we observe that the front of a car is often translated into a bird tail since they are both the thinner side of the objects. 
In this section, we solve this challenge in Bird $\leftrightarrow$ Car by exploring semi-supervised learning with a small amount of labeled data.

We labeled the heads/tails (headlights/rears) in 40 bird images and 40 car images (about $1.7\%$ of training images) for supervision as illustrated in Fig.~\ref{fig:semi}(a). The workload is not heavy: We just click the mouse at the head or tail position, where a circle with a fixed radius is drawn, resulting in a two-channel label map.
Then, we incorporate the high-level semantic prior from the pre-trained VGG network into GP-UNIT in two ways:
1) The first layer of our generator receives a concatenation of $E_c(x)$ and the $l$-layer VGG feature $\Phi_l(x)$ instead of only $E_c(x)$ as inputs;
2) We train networks to predict the label map of our labeled images based on their VGG features and add a loss to penalize the head/tail position changes during translation.

\begin{figure*}[t]
\centering
\includegraphics[width=1\linewidth]{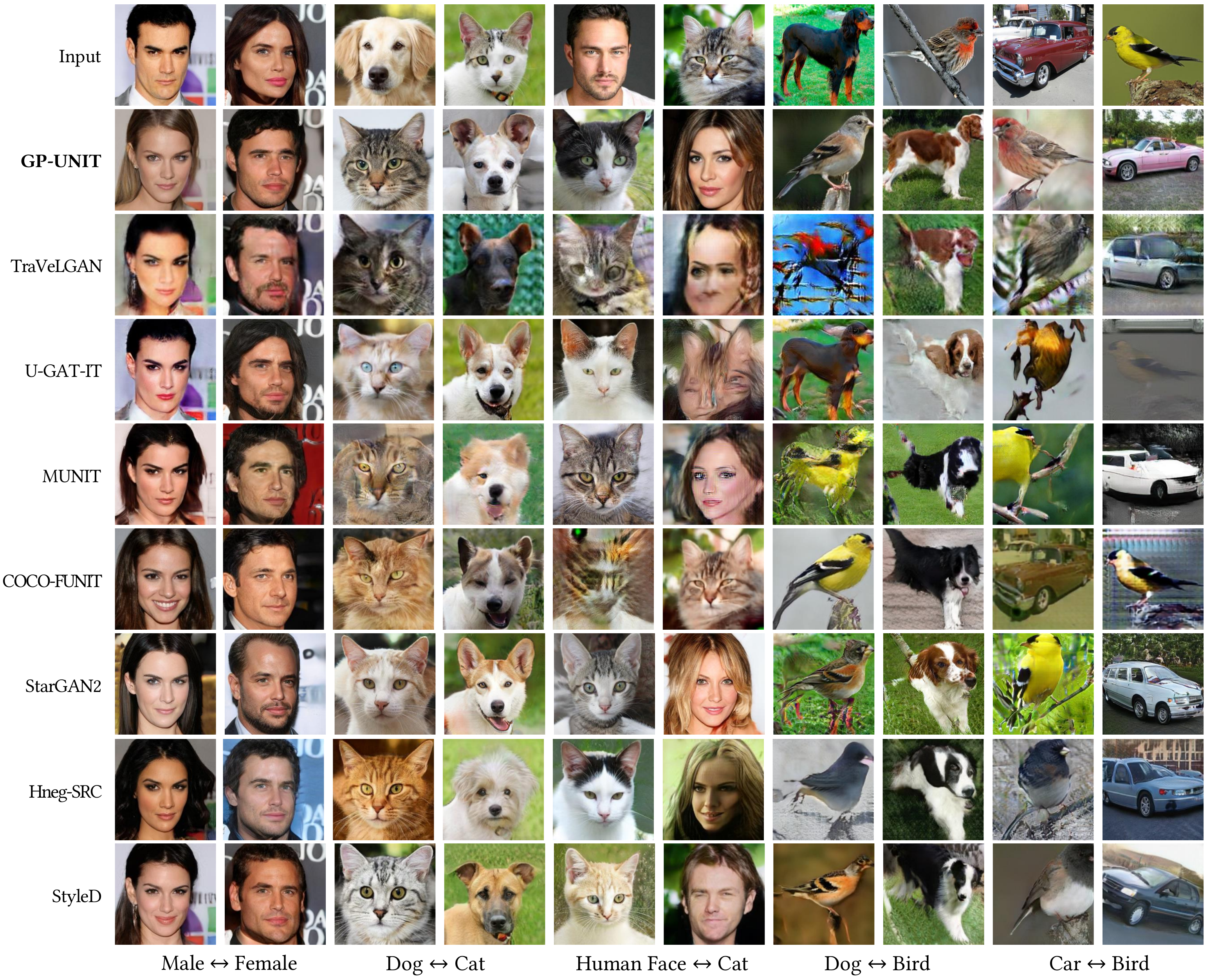}\vspace{-2mm}
\caption{Visual comparison with TraVeLGAN~\cite{amodio2019travelgan}, U-GAT-IT~\cite{kim2019u}, MUNIT~\cite{huang2018multimodal}, COCO-FUNIT~\cite{saito2020coco}, StarGAN2~\cite{choi2020stargan}, Hneg-SRC~\cite{jung2022exploring} and StyleD~\cite{kim2022style}. GP-UNIT consistently outperforms on all tasks and demonstrates greater superiority as the task becomes more challenging (from left to right).}
\vspace{-1mm}
\label{fig:comparison}
\end{figure*}

Specifically, for two domains $\mathcal{X}$ and $\mathcal{Y}$, we train two corresponding prediction networks $P_\mathcal{X}$ and $P_\mathcal{Y}$, with each composed of a convolutional layer and a Sigmoid layer. They map a concatenation of the VGG relu4\_2 feature and the upsampled relu5\_2 feature of an image to its label map.  $P_\mathcal{X}$ is trained with $L_2$ loss and KL-divergence loss between the ground truth label map $L_x$ of $x$ and the predict label map
$P_\mathcal{X}([\Phi_{\text{relu4\_2}}(x), \uparrow\Phi_{\text{relu5\_2}}(x)])$ (for simplicity, we will use $P_\mathcal{X}(x)$ to denote it in the following):
\begin{equation}
   \mathcal{L}_{label}=\mathbb{E}_{x}[\lambda_P\|P_\mathcal{X}(x)-L_x\|_2^2+D_{\text{KL}}(P_\mathcal{X}(x) \| L_x)],\\
\end{equation}
where $\lambda_P=250$. $P_\mathcal{Y}$ is similarly trained on the labeled images of $\mathcal{Y}$. The performance of $P_\mathcal{X}$ and $P_\mathcal{Y}$ on the training data and testing data is shown in Figs.~\ref{fig:semi}(a)(b).

Given the trained $P_\mathcal{X}$ and $P_\mathcal{Y}$, we penalize the head/tail position changes by comparing $P_\mathcal{X}(x)$ and $P_\mathcal{Y}(\hat{y})$. However, the predicted label map is not accurate enough and it is too strict to require exactly the same position, \eg, requiring the bird's head to be below its tail like a car. To solve this problem, we propose a robust position loss $\mathcal{L}_{pos}$ based on the centroid, which is intuitively illustrated in the last image of Fig.~\ref{fig:semi}(c).
We use the predicted value at each position as the mass to calculate the centroid of $P_\mathcal{X}(x)$ and $P_\mathcal{Y}(\hat{y})$, resulting in two sets of centroids $c_x$ and $c_{\hat{y}}$. Each set contains the coordinates of object's head and tail. Let the superscript $i$ ad $j$ denote the abscissa and the ordinate of the coordinates, respectively. $\mathcal{L}_{pos}$  is defined as
\begin{equation}
   \mathcal{L}_{pos}=\mathbb{E}_{x,y}\Big[\sum_{\gamma\in\{i,j\}}\lambda^\gamma_{p}\min(\|c^\gamma_x-c^\gamma_{\hat{y}}\|_2^2, \tau)\Big],\\
\end{equation}
where we care more about the semantic consistency in the horizontal direction by setting $\lambda^i_{p}=1$ and $\lambda^j_{p}=0.1$. The margin parameter $\tau=16$ is to prevent large penalties that possibly come from failure predictions.

With such simple supervision, our method finds better correspondences as in Fig.~\ref{fig:semi}(c).

\begin{table*} [t]
\caption{Quantitative comparison. We use FID and Diversity with LPIPS to evaluate the quality and diversity of the generated images.}\vspace{-1mm}
\label{tb:fid}
\centering
%\scriptsize
\begin{tabular*}{\textwidth}{@{\extracolsep{\fill}}l|c|c|c|c|c|c|c|c|c|c|c|c}
\toprule
Task & \multicolumn{2}{c|}{Male $\leftrightarrow$ Female} &  \multicolumn{2}{c|}{Dog $\leftrightarrow$ Cat} & \multicolumn{2}{c|}{Human Face $\leftrightarrow$ Cat}  &  \multicolumn{2}{c|}{Bird $\leftrightarrow$ Dog} &  \multicolumn{2}{c|}{Bird $\leftrightarrow$ Car} &   \multicolumn{2}{c}{Average} \\
\midrule
Metric & FID & Diversity & FID & Diversity & FID & Diversity & FID & Diversity & FID & Diversity & FID & Diversity \\
\midrule
TraVeLGAN & 66.60  & $-$ & 58.91  & $-$ & 85.28  & $-$ & 169.98  & $-$ & 164.28  & $-$ & 109.01  & $-$ \\
U-GAT-IT & 29.47  & $-$ & 38.31  & $-$ & 110.57  & $-$ & 178.23  & $-$ & 194.05  & $-$ & 110.12  & $-$ \\
\midrule
MUNIT & 22.64  & 0.37  & 80.93  & 0.47  & 56.89  & \textbf{0.53}  & 217.68  & 0.57  & 121.02  & 0.60  & 99.83  & 0.51 \\
COCO-FUNIT & 39.19  & 0.35  & 97.08  & 0.08  & 236.90  & 0.33  & 30.27  & 0.51  & 207.92  & 0.12  & 122.27  & 0.28 \\
StarGAN2 & \textbf{14.61}  & \textbf{0.45}  & 22.08  & 0.45  & \textbf{11.35}  & 0.51  & 20.54  & 0.52  & 29.28  & 0.58  & 19.57  & 0.50 \\
Hneg-SRC & 22.77 & 	0.25 & 	26.21 & 0.42 & 19.77 & 0.41 & 52.56 & 0.53 & 43.19 & 0.49 & 32.90 & 0.42 \\
StyleD & 15.20 & 0.26 & 32.08 & 0.42 & 26.03 & 0.35 & 22.71 & 0.58 & 48.35 & 0.59 & 28.87 & 0.44 \\
\textbf{GP-UNIT} & 14.63  & 0.37  & \textbf{15.29}  & \textbf{0.51}  & 13.04  & 0.49  & \textbf{11.29}  & \textbf{0.60}  & \textbf{13.93}  & \textbf{0.61}  & \textbf{13.64}  & \textbf{0.52} \\
\bottomrule
\end{tabular*}
\end{table*}

\begin{figure*}[t]
\centering
\includegraphics[width=1\linewidth]{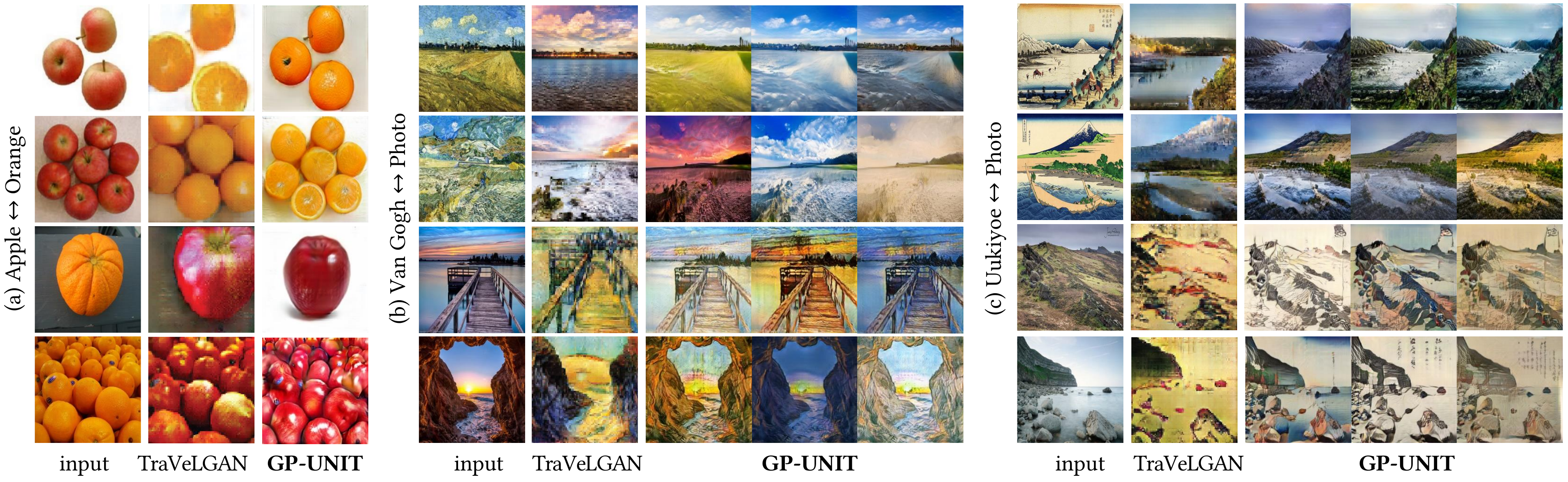}\vspace{-2mm}
\caption{Comparison with TraVeLGAN\protect\footnotemark. GP-UNIT is comparable to TraVeLGAN in visual quality and surpasses it in image resolution and diversity.}\vspace{-1mm}
\label{fig:travelgan}
\end{figure*}

\section{Experimental Results}
\label{sec:experiment}

\noindent
\textbf{Dataset.} In the first stage, we prepare both synthesized data and real data.
For synthesized data, we use the official BigGAN
\cite{brock2018large} to generate correlated images associated by random latent codes for each of the 291 domains including animals and vehicles. After filtering low-quality ones, we finally obtain 655 images per domain that are linked across all domains, 600 of which are for training. We denote this dataset as \textit{synImageNet-291}.
For real data, we apply HTC~\cite{chen2019hybrid} to ImageNet~\cite{russakovsky2015imagenet} to detect and crop the object regions. Each domain uses 600 images for training. We denote this dataset as \textit{ImageNet-291}.
Besides, 29K face images of CelebA-HQ~\cite{liu2015deep, karras2018progressive} are also included for training.

In the second stage, we evaluate our standard GP-UNIT on four translation tasks.~1) Male $\leftrightarrow$ Female on 28K training images of  CelebA-HQ~\cite{liu2015deep, karras2018progressive}.
2) Dog $\leftrightarrow$ Cat on AFHQ~\cite{choi2020stargan}, with 4K training images per domain.
3) Human Face $\leftrightarrow$ Cat on 4K AFHQ images and 29K CelebA-HQ images.
4) Bird $\leftrightarrow$ Dog or Car:
Four classes of birds, four classes of dogs
and four classes of cars
in \textit{ImageNet-291} are used. Every four classes form a domain with 2.4K training images.
Here, Bird$\leftrightarrow$ Car is used as the extreme case to test to what extent GP-UNIT can handle for stress testing.
We perform evaluation on our controllable GP-UNIT on Male $\leftrightarrow$ Female and Dog $\leftrightarrow$ Cat, and show our results with semi-supervised learning on the challenging Bird $\leftrightarrow$ Car.

\noindent
\textbf{Network training.} We set $\lambda_{s}=5$, $\lambda_{r}=0.001$, $\lambda_{1}=\lambda_{3}=\lambda_{4}=1$ and $\lambda_{2}=50$. For semi-supervised translation in Sec.~\ref{sec:semi}, we lower $\lambda_{1}$ to $0.1$.
For Cat $\rightarrow$ Human Face, we use an additional identity loss~\cite{deng2019arcface} with weight $1$ to preserve the identity of the reference face following~\cite{richardson2020encoding}.
Dynamic skip connections are applied to the 2nd layer ($l=1$) and the 3rd layer ($l=2$) of $G$.
Except for Male $\leftrightarrow$ Female, we do not use dynamic skip connections to compute $\mathcal{L}_{rec}$.
During training the controllable GP-UNIT, for our six-layer $E_c$, we set $[\lambda_{cc}^0, ..., \lambda_{cc}^5]$ in Eq.~(\ref{eq:cc1}) to $[1,1,0,0,0,0]$ for Cat $\rightarrow$ Dog and $[1,1,1,0,0,0]$ for Dog $\rightarrow$ Cat, respectively. We empirically set $\lambda_{cc}$ in Eq.~(\ref{eq:cc2}) to $0.25$ and $0.3$ for Male $\rightarrow$ Female and Female $\rightarrow$ Male, respectively.

\footnotetext{TraVeLGAN's results are directly taken from its paper. TraVeLGAN crops the input images, therefore the scales of its results do not match the original input images we used.}

\subsection{Comparison with the State of the Arts}

\begin{figure*}[t]
\centering
\includegraphics[width=1\linewidth]{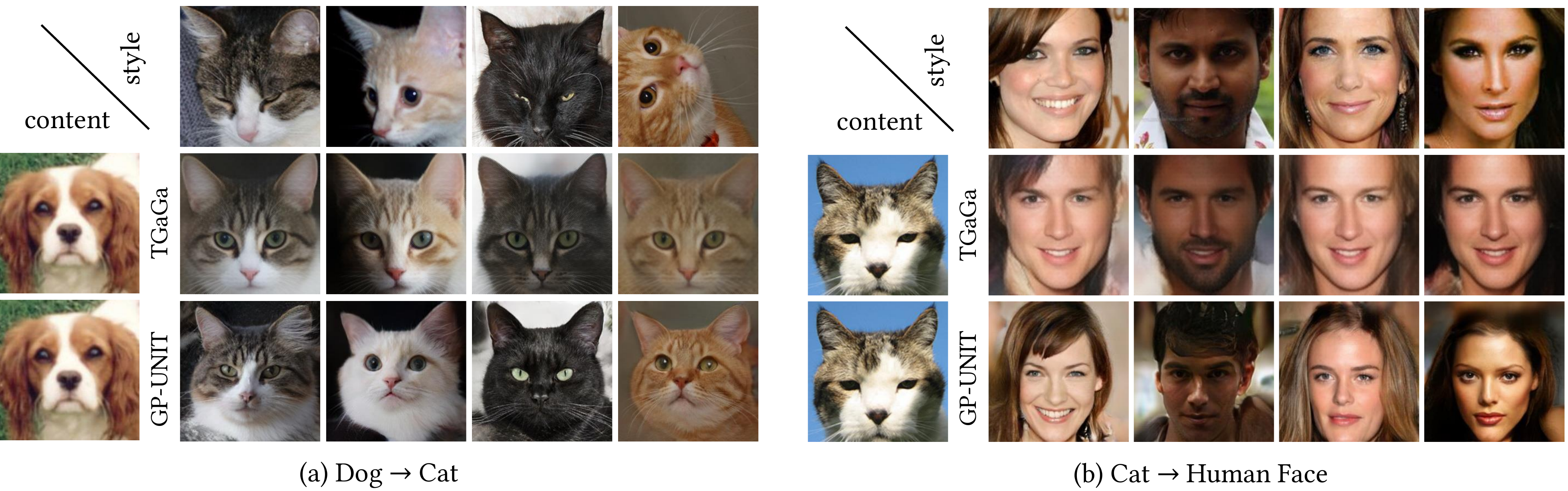}\vspace{-2mm}
\caption{Comparing exemplar-guided translation with TGaGa\protect\footnotemark. GP-UNIT surpasses TGaGa in vivid details and style consistency.}\vspace{-1mm}
\label{fig:transgaga}
\end{figure*}

\noindent
\textbf{Qualitative comparison.} We perform visual comparison to eight state-of-the-art methods in Fig.~\ref{fig:comparison} and Fig.~\ref{fig:transgaga}.
As shown in Fig.~\ref{fig:comparison}, cycle-consistency-guided U-GAT-IT~\cite{kim2019u}, MUNIT~\cite{huang2018multimodal} and StarGAN2~\cite{choi2020stargan} rely on the low-level cues of the input image for bi-directional reconstruction, which leads to some undesired artifacts, such as the distorted cat face region that corresponds to the dog ears, and the ghosting dog legs in the generated bird images.
Meanwhile, TraVeLGAN~\cite{amodio2019travelgan} and COCO-FUNIT~\cite{saito2020coco} fail to build proper content correspondences for Human Face $\leftrightarrow$ Cat and Bird $\leftrightarrow$ Car.
Hneg-SRC~\cite{jung2022exploring} explores patch-level contrastive learning, which might fail to capture large-scale shape correspondences, thus distorting the animal shapes for Bird $\leftrightarrow$ Dog.
StyleD~\cite{kim2022style} generates plausible results, but is less robust to unrelated details like the cat collar, and its synthesized bird has less correspondence to the input car image.
By comparison, our method is comparable to the above methods on Male $\leftrightarrow$ Female task and show consistent superiority on other challenging tasks.
In Fig.~\ref{fig:travelgan}, we compare our model to TraVeLGAN on domains that are not  observed by the content encoder in Stage I.
The 128$\times$128 results of TraVeLGAN have less details than our 256$\times$256  results. In addition, our method supports multi-modal translation to improve the diversity of the results.
In Fig.~\ref{fig:transgaga}, we compare our model to TGaGa~\cite{wu2019transgaga}, which also deals with large geometric deformations
on exemplar-guided translation.
The results of TGaGa are blurry. Moreover, TGaGa fails to match the example appearance precisely. For example, the generated cats and human faces share very similar shapes and textures, except for the color changes. Our GP-UNIT surpasses TGaGa in both vivid details and style consistency.

\begin{table*} [t]
\caption{User preference scores in terms of content consistency (CC) and overall preference (OP). Best scores are marked in bold.}
\label{tb:user_study}
\centering
\begin{tabular*}{\textwidth}{@{\extracolsep{\fill}}l|c|c|c|c|c|c|c|c|c|c|c|c}
\toprule
Task & \multicolumn{2}{c|}{Male $\leftrightarrow$ Female} &  \multicolumn{2}{c|}{Dog $\leftrightarrow$ Cat} & \multicolumn{2}{c|}{Human Face $\leftrightarrow$ Cat}  &  \multicolumn{2}{c|}{Bird $\leftrightarrow$ Dog} &  \multicolumn{2}{c|}{Bird $\leftrightarrow$ Car} &   \multicolumn{2}{c}{Average} \\
\midrule
Metric & CC & OP & CC & OP & CC & OP & CC & OP & CC & OP & CC & OP \\
\midrule
TraVeLGAN & 0.015  & 0.000  & 0.020  & 0.015  & 0.021  & 0.023  & 0.021  & 0.008  & 0.012  & 0.015  & 0.018  & 0.012 \\
UGATIT & 0.158  & 0.114  & 0.102  & 0.054  & 0.031  & 0.024  & 0.036  & 0.008  & 0.025  & 0.008  & 0.070  & 0.042 \\
MUNIT & 0.065  & 0.079  & 0.034  & 0.024  & 0.061  & 0.040  & 0.028  & 0.015  & 0.019  & 0.032  & 0.041  & 0.038 \\
COCOUNIT & 0.081  & 0.106  & 0.084  & 0.079  & 0.056  & 0.000  & 0.075  & 0.050  & 0.058  & 0.008  & 0.071  & 0.049 \\
StarGAN & 0.088  & 0.098  & 0.206  & 0.186  & 0.090  & 0.097  & 0.137  & 0.138  & 0.097  & 0.086  & 0.124  & 0.121 \\
Hner-SRC & \textbf{0.213}  & 0.190  & 0.073  & 0.066  & 0.140  & 0.146  & 0.032  & 0.037  & 0.020  & 0.013  & 0.096  & 0.090 \\
StyleD & 0.169  & 0.187  & 0.159  & 0.213  & 0.143  & 0.166  & 0.065  & 0.072  & 0.165  & 0.178  & 0.140  & 0.163 \\
\textbf{GP-UNIT} & 0.211  & \textbf{0.226}  & \textbf{0.322} & \textbf{0.363}  & \textbf{0.458}  & \textbf{0.504}  & \textbf{0.606} & \textbf{0.672}  & \textbf{0.604}  & \textbf{0.660}  & \textbf{0.440}  & \textbf{0.485} \\
\bottomrule
\end{tabular*}\vspace{-1mm}
\end{table*}

\begin{figure*}[t]
\centering
\includegraphics[width=0.98\linewidth]{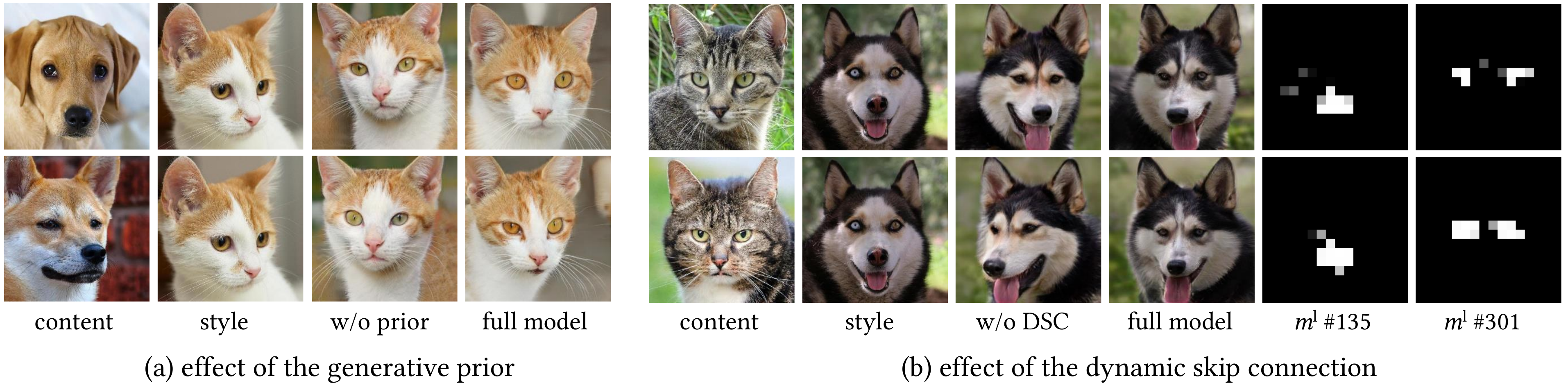}\vspace{-2mm}
\caption{Ablation study on the generative prior and the dynamic skip connection.}\vspace{-1mm}
\label{fig:ablation}
\end{figure*}

\noindent
\textbf{Quantitative comparison.}~We follow~\cite{wu2019transgaga,choi2020stargan} to perform quantitative comparison  in terms of quality and diversity.
FID~\cite{heusel2017gans} and LPIPS~\cite{zhang2018unreasonable} are used to evaluate the image quality of generated results against the real data and the output diversity, respectively.
For methods supporting multi-modal transfer (MUNIT, COCO-FUNIT, StarGAN2, Hneg-SRC, StyleD, GP-UNIT), we generate 10 paired translation results per test image from randomly sampled latent codes or exemplar images to compute their average LPIPS distance.
The quantitative results averaged over all test images are reported in Table~\ref{tb:fid}, which
are consistent with Fig.~\ref{fig:comparison}, \ie, our method is comparable or superior to the compared methods, and the advantage becomes more distinct on difficult tasks, obtaining the best overall FID and diversity.
We find GP-UNIT and StyleD tend to preserve the background of the input image. This property does not favor diversity, but might be useful in some applications. On close domains, StyleD even keeps the background unchanged, and it samples the style based on a few learned prototypes, which harms the diversity, leading to its high FID and low LPIPS. The shape distortions greatly harm the FID score of Hneg-SRC on Bird $\leftrightarrow$ Dog.
Although StarGAN2 yields realistic human faces  (best FID) on Cat $\rightarrow$ Human Face, it ignores the pose correspondences with the input cat faces (lower content consistency than GP-UNIT), as in Fig.~\ref{fig:comparison}.

\footnotetext{At the time of this submission, the code and training data of TGaGa are not released. We directly use the test and result images kindly provided by the authors of TGaGa. Since the training data of GP-UNIT and TGaGa does not match, this comparison is for visual reference only.}

\begin{figure*}[t]
\centering
\includegraphics[width=0.98\linewidth]{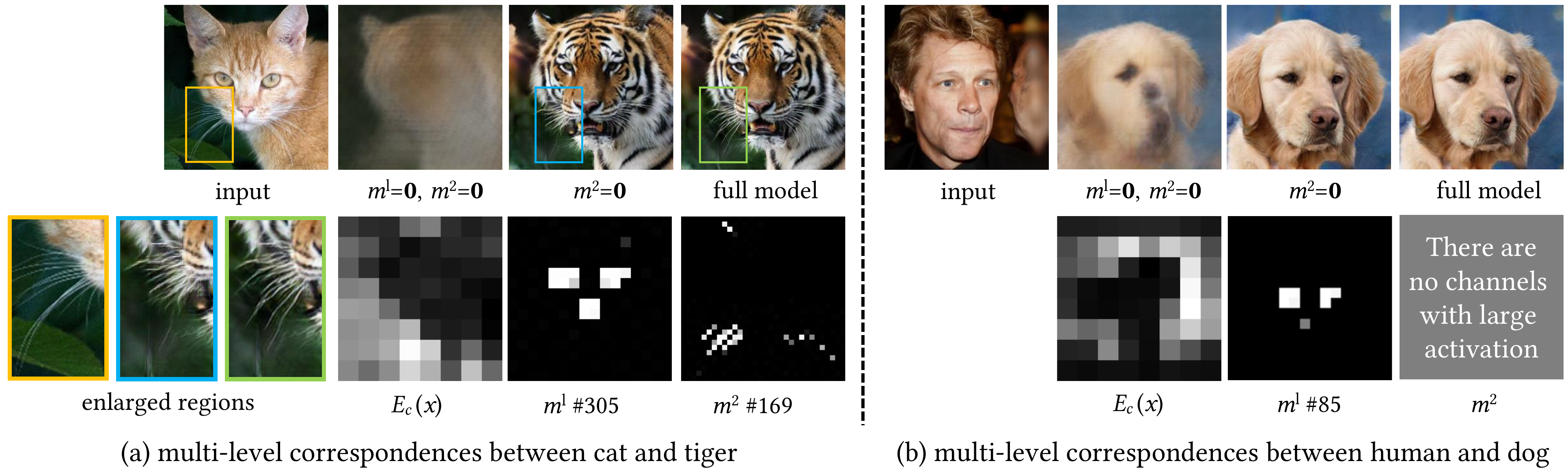}\vspace{-2mm}
\caption{Our framework learns multi-level content correspondences that are robust and adaptable to different translation tasks.}\vspace{-1mm}
\label{fig:correspondence}
\end{figure*}

We further conduct a user study to evaluate the content consistency and overall translation performance. A total of 13 subjects participate in this study to select what they consider to be the best results from the eight methods. Since in some tasks like Male $\leftrightarrow$ Female, the performance of each method is similar, we allow multiple selections. For each selection, if a user select results from $N$ methods as the best results, those methods get $1/N$ scores, and other methods get $0$ scores. A total of $1751$ selections on 50 groups of results (Every first five groups of Figs.~3-12 in the supplementary pdf) are tallied. Table~\ref{tb:user_study} summarizes the average user scores, where GP-UNIT receives notable preference for both content consistency and overall performance.

\subsection{Ablation Study}
\label{sec:ablation}

\noindent
\textbf{Generative prior distillation.} As shown in Fig.~\ref{fig:ablation}(a), if we train our content encoder from scratch along with all other subnetworks in the second stage, like most image translation frameworks, this variant fails to preserve the content features such as the eye position. By comparison, our pre-trained content encoder successfully exploits the generative prior to build effective content mappings. It also suggests the necessity of the coarse-level content feature, only based on which valid finer-level features can be learned. Hence, the generative prior is the key to the success of our coarse-to-fine scheme of content correspondence learning.

\noindent
\textbf{Dynamic skip connection.} As shown in Fig.~\ref{fig:ablation}(b), without dynamic skip connections (DSC), the model cannot keep the relative position of the nose and eyes as in the content images. We show that the 135th and 301st channels of the mask $m^1$ predicted by our full model effectively locate these features for accurate content correspondences.
To better understand the effect of the generative prior and DSC, we perform quantitative comparison in terms of quality, diversity and content consistency. For content consistency, ten users are invited to select the best one from the results of three configurations in terms of content consistency. FID, Diversity averaged over the whole testing set and the user score averaged over six groups of results are presented in Table~\ref{tb:user_study2}. Out full model achieves the best FID and content consistency. The three configurations generate images with comparable diversities. Since the generative prior and dynamic skip connection provide content constraints, as expected, results of the model without them are more diverse.

\begin{table} [t]
\caption{Ablation study in terms of FID, Diversity and content consistency (CC).}
\label{tb:user_study2}
\centering
\begin{tabular}{l|c|c|c}
\toprule
Metric & FID & Diversity & CC \\
\midrule
w/o generative prior	& 16.11	& \textbf{0.55}	& 0.02 \\
w/o dynamic skip connection	& 15.83	& 0.52	& 0.15 \\
full model	& \textbf{15.29}	& 0.51	& \textbf{0.83} \\
\bottomrule
\end{tabular}
\end{table}

\noindent
\textbf{Multi-level cross-domain correspondences.}~Figure~\ref{fig:correspondence} analyzes the learned multi-level correspondences. The most abstract $E_c(x)$ only gives layout cues. If we solely use $E_c(x)$ (by setting both masks $m^1$ and $m^2$ to all-zero tensors), the resulting tiger and dog faces have no details. Meanwhile, $m^1$ focuses on mid-level details like the nose and eyes of cat face in the 305th channels, and eyes of human face in the 85the channels, which is enough to generate a realistic result with $E_c(x)$. Finally, $m^2$ pays attention to subtle details like the cat whiskers in the 169th channel for close domains.
Therefore, our full multi-level content features enable us to simulate the extremely fine-level long whiskers in the input. As expected, such kind of fine-level correspondences are not found between more distant human and dog faces, preventing the unwanted appearance influence from the source domain (\eg, clothes in the generated cat in Fig.~\ref{fig:comparison}).

\begin{figure}[t]
\centering
\includegraphics[width=0.98\linewidth]{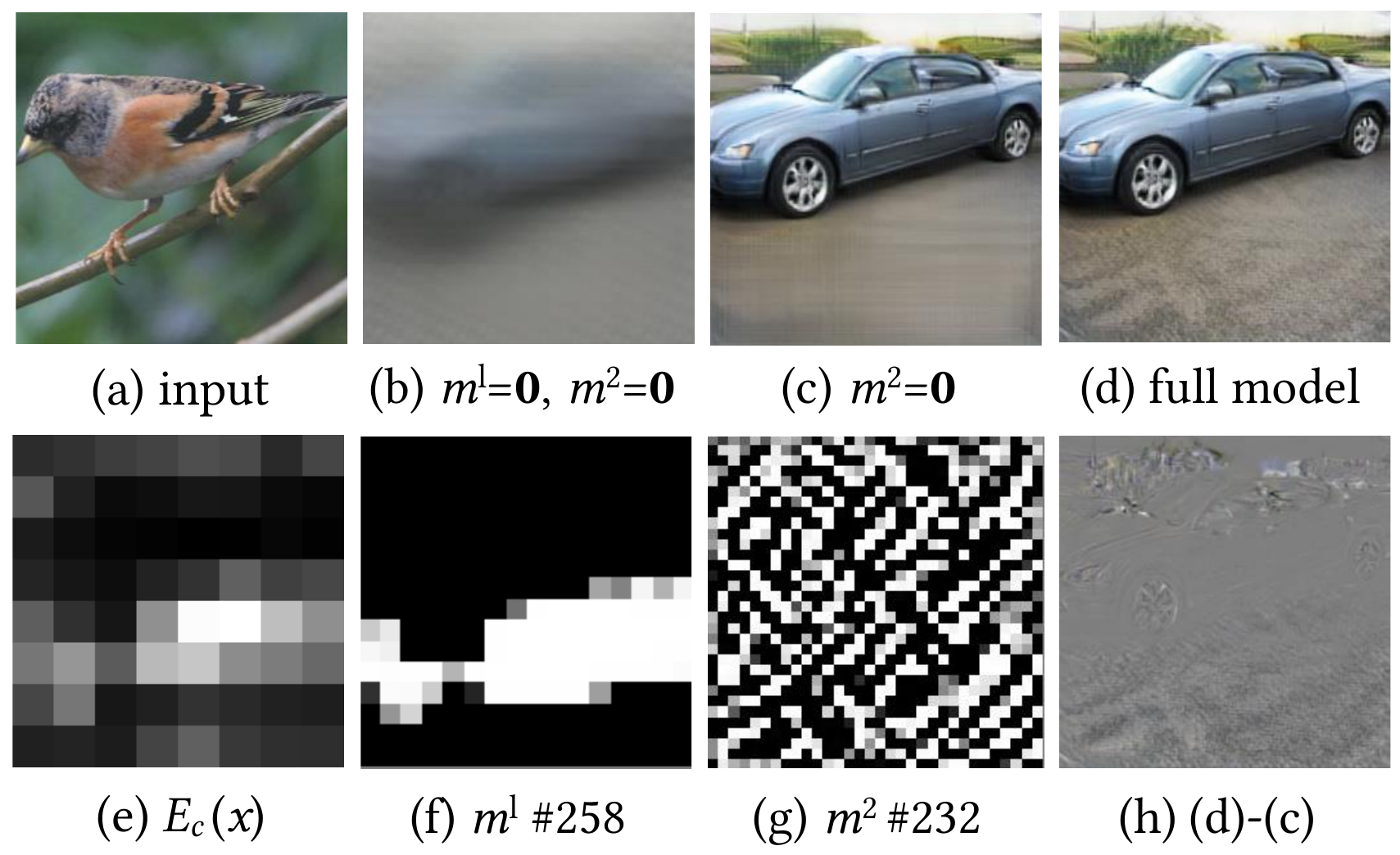}\vspace{-2mm}
\caption{Learned multi-level content correspondences for Bird $\rightarrow$ Car.}\vspace{-1mm}
\label{fig:correspondence2}
\end{figure}

In Fig.~\ref{fig:correspondence2}, we analyze the content features between distant domains. $E_c(x)$ can solely generate a rough layout of the car. The meaning of $m^1$ is not as intuitive as other tasks, since birds and cars have almost no semantic relationship. However, we can still find some reasonable cues, \eg, $m^1$ focuses on the background areas around the foreground object in the 258th channel. Finally, the 232nd channel of $m^2$ extracts high-frequency signals (similar to isophote map) from the input to add texture details in the background, as indicated by the difference map in Fig.~\ref{fig:correspondence2}(h). Such behavior reduces the learning difficulty of the generator $G$, allowing $G$ to better focus on the structure synthesis. In Fig.~\ref{fig:correspondence}, we do not find such behavior, likely due to the fact that the backgrounds in AFHQ~\cite{choi2020stargan} and CelebA-HQ~\cite{karras2018progressive} are mostly blurry with few high-frequency details.

Note that the aforementioned reasonable and adaptable semantic attentions are learned merely via the generation prior, without any explicit correspondence supervision.

\begin{figure}[t]
\centering
    \includegraphics[width=0.96\linewidth]{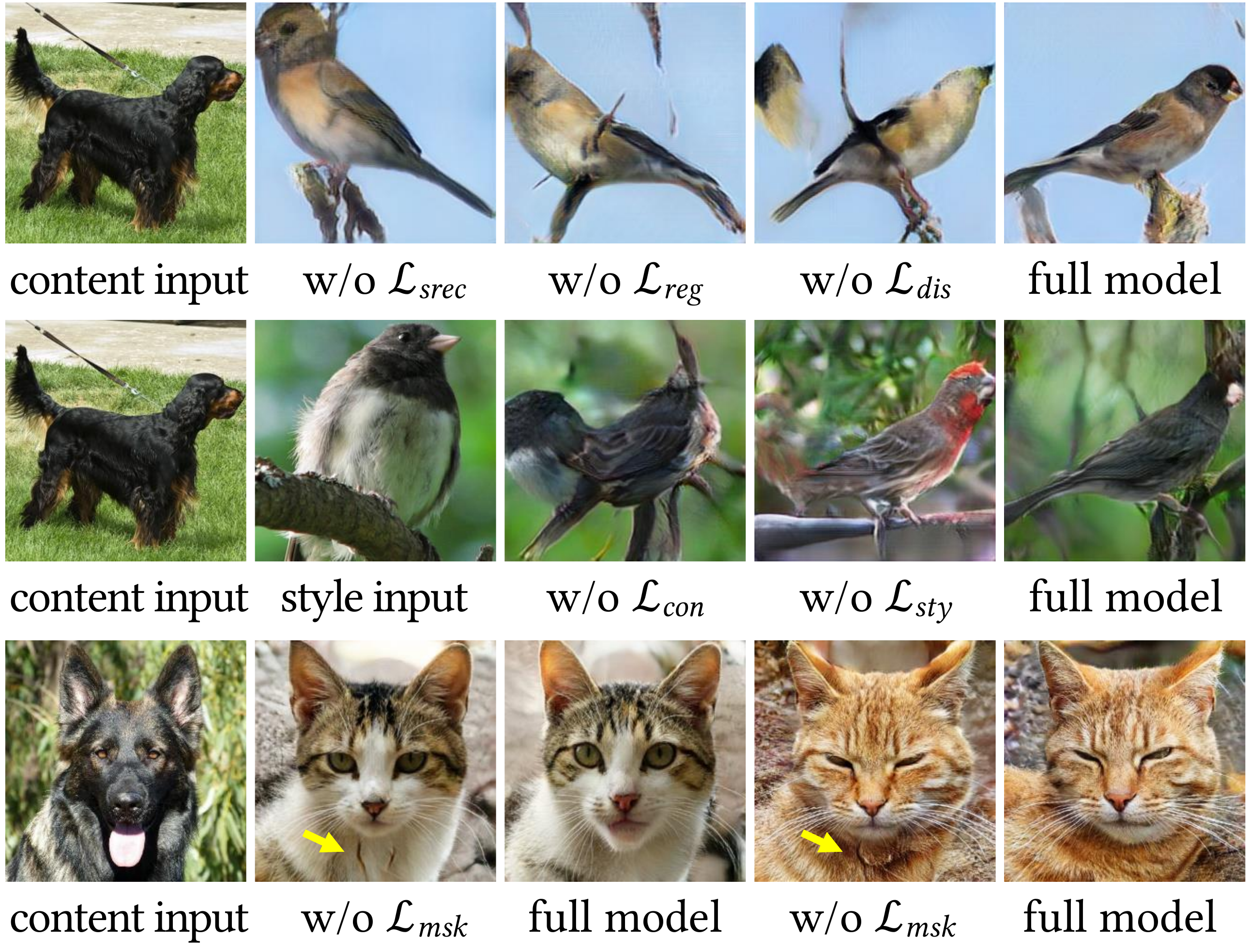}\vspace{-2mm}
\caption{Effects of the loss terms.}\vspace{-1mm}
\label{fig:ablation0}
\end{figure}

\noindent
\textbf{Loss functions.} Figure~\ref{fig:ablation0} studies the effects of the loss terms.
In Stage I, $\mathcal{L}_{srec}$ is the key to learn correct content features, or correspondence is not built.
$\mathcal{L}_{reg}$ makes content features sparser to improve robustness to unimportant domain-specific details.
$\mathcal{L}_{dis}$ finds domain-shared features to prevent the output from affected by objects from the source domain like the dog tail.
In Stage II, $\mathcal{L}_{con}$ helps strengthen the pose correspondence while $\mathcal{L}_{sty}$ makes the output better match the style of the  exemplar image.
Without $\mathcal{L}_{msk}$, it can be observed that the structure of the dog's tongue appears in the translation results as indicated by the yellow arrows. $\mathcal{L}_{msk}$ effectively imposes sparsity over the mask to prevent overuse of the content information.

\subsection{Control Content Consistency}
\label{sec:exp_control}

\begin{figure}[t]
\centering
\includegraphics[width=1\linewidth]{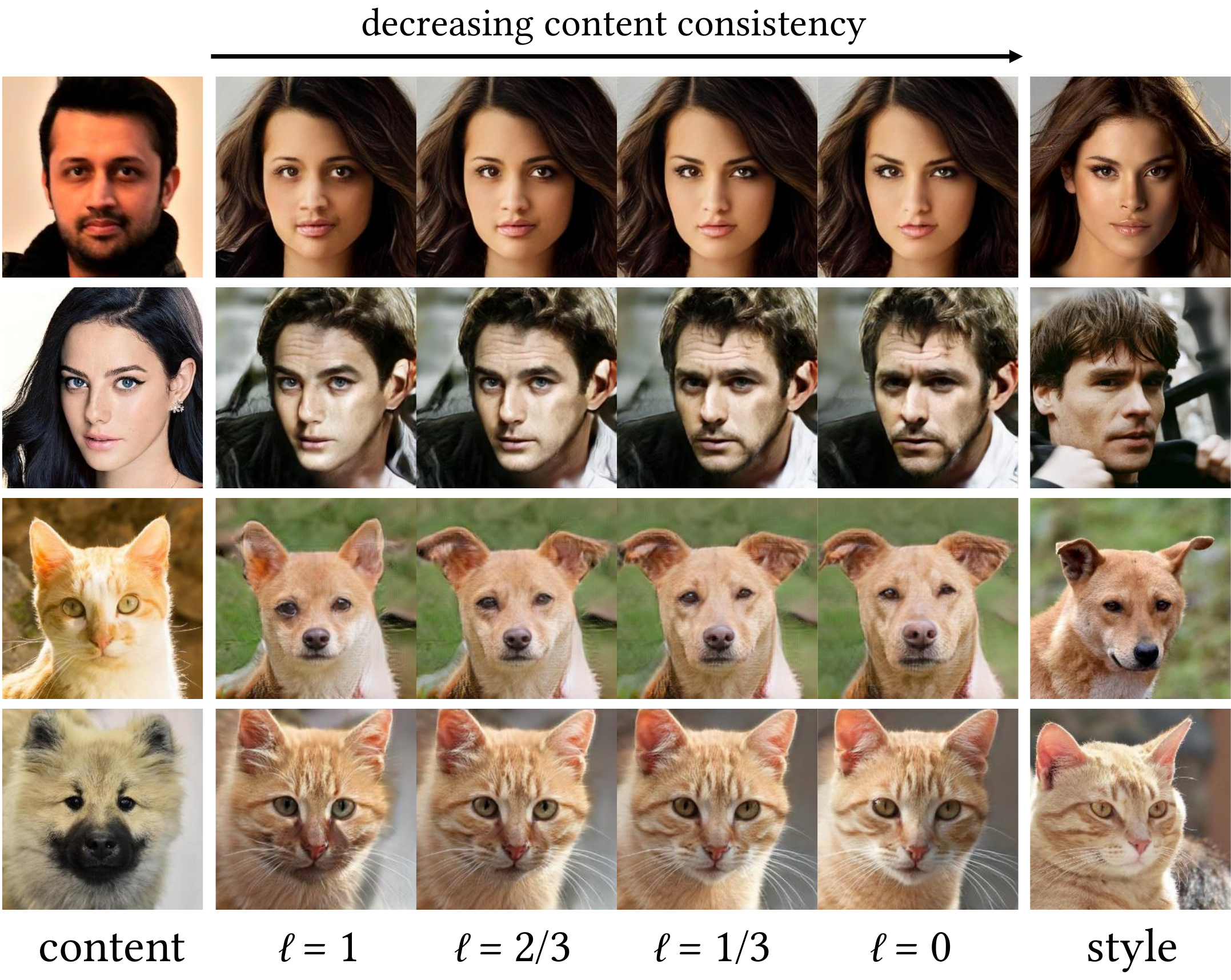}\vspace{-2mm}
\caption{The proposed controllable version of GP-UNIT provides a parameter $\ell$ to adjust the content consistency during translation.}\vspace{-1mm}
\label{fig:exp_control}
\end{figure}

\begin{figure*}[t]
\centering
\includegraphics[width=0.98\linewidth]{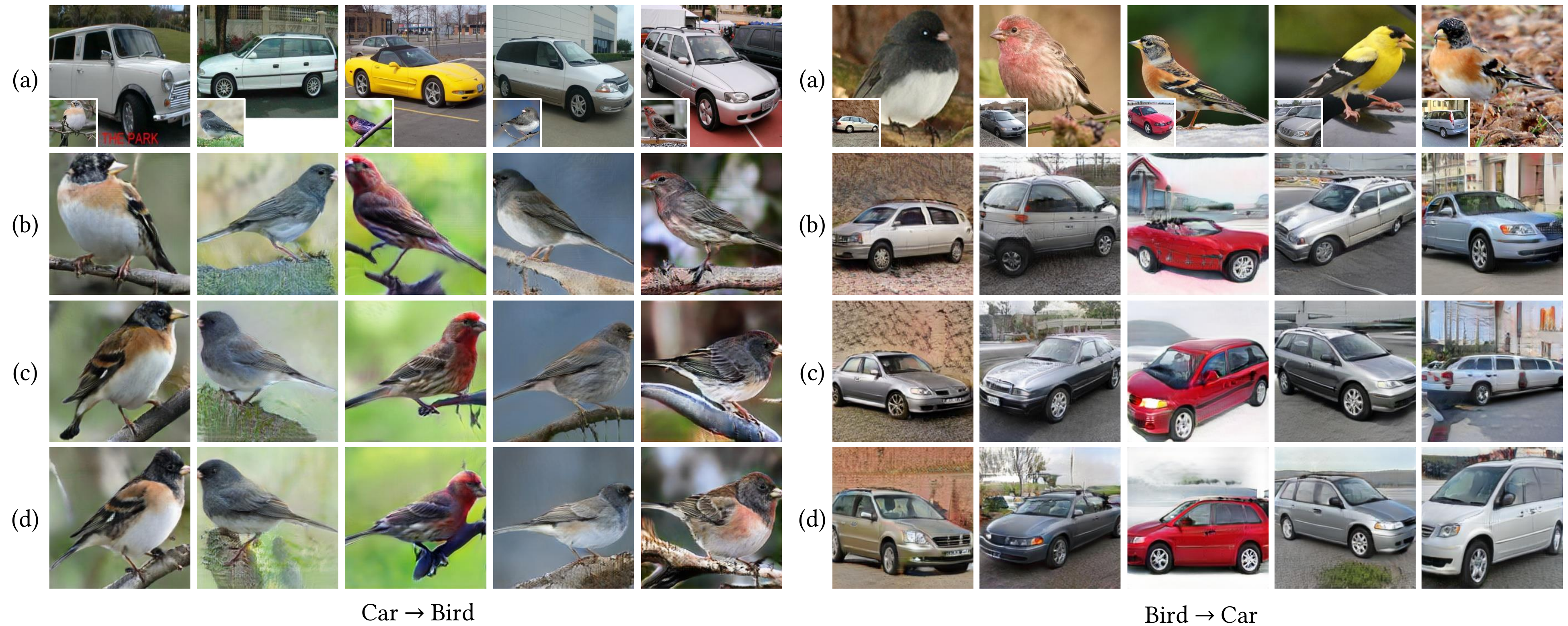}\vspace{-2mm}
\caption{By exploring semi-supervised learning with a small amount of labeled data, GP-UNIT successfully builds valid semantic correspondences between the bird heads and the car fronts, which is hard to learn solely from the objects' appearance. (a) Input content image and style image. (b) Results of the standard GP-UNIT without semi-supervised learning. (c) Results with semi-supervised learning on 4 labeled data pairs. (d) Results with semi-supervised learning on 40 labeled data pairs.}\vspace{-2mm}
\label{fig:compare_semi}
\end{figure*}

This section studies the effects of the parameter $\ell$  on adjusting the content consistency. In Fig.~\ref{fig:exp_control}, as $\ell$ uniformly increases, the translation results present more facial features from the content image, such as the human identities, the shape of the ears and the position of the eyes.
Correspondingly, the style of the results and the style images match sparingly, \eg, the white fur around the mouth turns dark in the Dog $\rightarrow$ Cat case. As can be observed, there is a trade-off between the style consistency and content consistency. Our method provides users with a flexible solution, allowing them to adjust $\ell$ freely to find the most ideal state.

\begin{table} [t]
\caption{Effect of $\ell$ on the style consistency and content consistency.}\vspace{-1mm}
\label{tb:control}
\centering
%\scriptsize
\begin{tabular}{l|c|c|c|c}
\toprule
Task & \multicolumn{2}{c|}{Male $\leftrightarrow$ Female} & \multicolumn{2}{c}{Dog $\leftrightarrow$ Cat} \\
\midrule
Metric & Style  & Content  & Style  & Content  \\
\midrule
$l=0.00$ & 0.146  & \textbf{0.934} & 0.124 & \textbf{0.845}\\
$l=0.25$  & \textbf{0.143}  & 0.985 & 0.119 & 0.960\\
$l=0.50$  &  \textbf{0.143}  & 1.062 & 0.114 & 1.062\\
$l=0.75$ &  \textbf{0.143}  & 1.129 & 0.111 & 1.143\\
$l=1.00$ &  0.145  & 1.171 & \textbf{0.109} & 1.192\\
\bottomrule
\end{tabular}%\vspace{-4mm}
\end{table}

\begin{figure*}[t]
\centering
\includegraphics[width=1\linewidth]{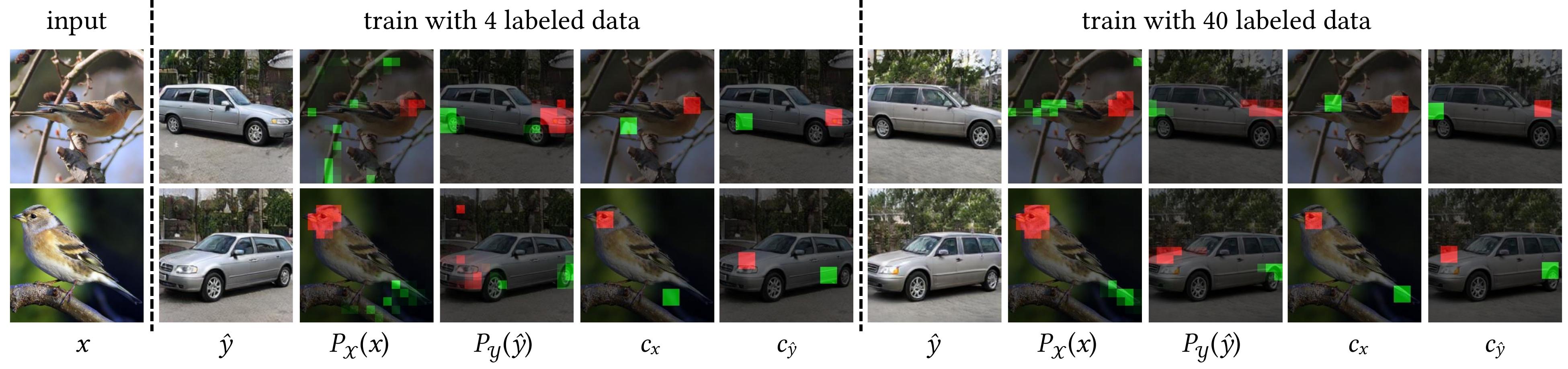}\vspace{-2mm}
\caption{Visual comparison of the label maps and centroids predicted by our methods trained with 4 and 40 labeled data.}\vspace{-2mm}
\label{fig:label_amount}
\end{figure*}

To quantitatively verifies the effect of the parameter $\ell$, we report the style consistency and content consistency over 500 paired Cat-Dog testing images and 1,000 paired Female-Male testing images in Table~\ref{tb:control}. Following the loss definition in AdaIN~\cite{huang2017adain}, we use the content loss between $\hat{y}$ and $x$, and the style loss between $\hat{y}$ and $y$ as the metric. Note that during training our method optimizes $\mathcal{L}_{sty}$ and $\mathcal{L}_{con}$ (Eqs.~(\ref{eq:content_loss})(\ref{eq:style_loss})) rather than AdaIN-based losses. Thus, the latter is more suitable as the evaluation metric.
It can be seen that $\ell$ does not affect the style much on human faces since the used style loss is insensitive to human identity, while $\ell$ and the content consistency are negatively correlated. On Dog $\leftrightarrow$ Cat, we can observe a clear trade-off between the style consistency and content consistency. Improving one factor will hinder another, and both are well controlled by $\ell$.
We further conduct a user study where we select the first 10 Cat-Dog testing image pairs and the first 10 Female-Male testing image pairs to generate 40 groups of results under $\ell=0$ and $\ell=1$, respectively. Then 15 users are asked to select one from the two results that is more content consistent to the content image and one that is more style consistent to the style image. The result shows that $85.7\%$ users agree that $\ell=0$ leads to better content consistency, and $81.3\%$ users agree that $\ell=1$ achieves better style consistency.

\subsection{Semi-Supervised Translation}
\label{sec:exp_semi}

In Fig.~\ref{fig:compare_semi}, we compare the performance of the standard GP-UNIT and the reinforced GP-UNIT trained with semi-supervision learning. The standard GP-UNIT builds correspondences by appearance, so that it maps the thinner and lower part of objects in two domains, \ie, bird tails and car fronts as in Fig.\ref{fig:compare_semi}(b), which is however counter-intuitive to humans. By providing a small amount of labeled data for supervision and taking use of the discriminative prior of VGG, GP-UNIT effectively find semantically more valid correspondences as humans as in Fig.~\ref{fig:compare_semi}(d). It suggests that translation benefits from both generative priors and discriminative priors of the popular deep models.

To investigate how the amount of labeled data impacts the performance, we conduct an ablation study where we train our model on 0, 4, 8, 12, 40 labeled image pairs. Then we perform translations on 100 bird and 100 car testing images. Finally, we manually count the proportion of the results where the head/headlight and tail/rear are located horizontally consistently to their content images  (\ie, if the bird head and tail are on the left and right in the content image, then the car headlight and rear in the translation result should also be on the left and right, respectively).
Table~\ref{tb:data_amount} reports the accuracy (proportion of the consistent results) of our methods with different training settings. It can be seen that only 4 labeled image pairs can increase the accuracy by $73.5\%$ and further increasing the amount of labeled data marginally improves the accuracy. Figure~\ref{fig:compare_semi}(c) qualitatively shows the results by our model trained with only 4 labeled image pairs where almost all semantic correspondences with the input images are correct.
We further conduct a user study where we perform translations on the first 10 bird and 10 car testing images with the model trained on 0, 4, 40 labeled data. Then 15 users are asked for each result whether the head/headlight and tail/rear are located consistently to the content image. According to the survey, users think that only $17.3\%$ results are consistent with no supervision, and the percentage increases to $69.0\%$ and further to $72.7\%$ with $4$ and $40$ labeled data, respectively.

The proposed position loss $\mathcal{L}_{pos}$ is the key to the robustness of our method to the limited data. Figure~\ref{fig:label_amount} visualizes the label maps and centroids predicted by our methods trained with 4 and 40 labeled data, respectively. It can be seen that the prediction network can only roughly locate the head/tail (headlight/rear) regions with limited training data. This issue is effectively solved by calculating the centroid, which aggregates rough regions into an accurate position for more robust supervision. If we directly penalize the changes of the predicted label maps with $L_2$ loss rather than $\mathcal{L}_{pos}$, the performance with 40 labeled data is even worse than that using $\mathcal{L}_{pos}$ with 4 labeled data as in Table~\ref{tb:data_amount}. It further verifies the importance of $\mathcal{L}_{pos}$.

\begin{table} [t]
\caption{Accuracy of different settings to keep consistent layouts.}\vspace{-1mm}
\label{tb:data_amount}
\centering
%\scriptsize
\begin{tabular}{l|c|c|c|c|c|c}
\toprule
Settings & 0  & 4  & 8  & 12 & 40-$L_2$ & 40  \\
\midrule
Bird$\rightarrow$ Car & 0.10  & 0.88 & \textbf{1.00} & \textbf{1.00} & 0.44 & \textbf{1.00} \\
Car$\rightarrow$ Bird & 0.22  & \textbf{0.91} & 0.80 & 0.76 & 0.05 & \textbf{0.92} \\
\midrule
Averaged & 0.160  & 0.895 & 0.900 & 0.880 & 0.245 & \textbf{0.960} \\
\bottomrule
\end{tabular}%\vspace{-4mm}
\end{table}

\subsection{More Results}
\label{sec:more_result}

\begin{figure}[t]
\centering
\includegraphics[width=1\linewidth]{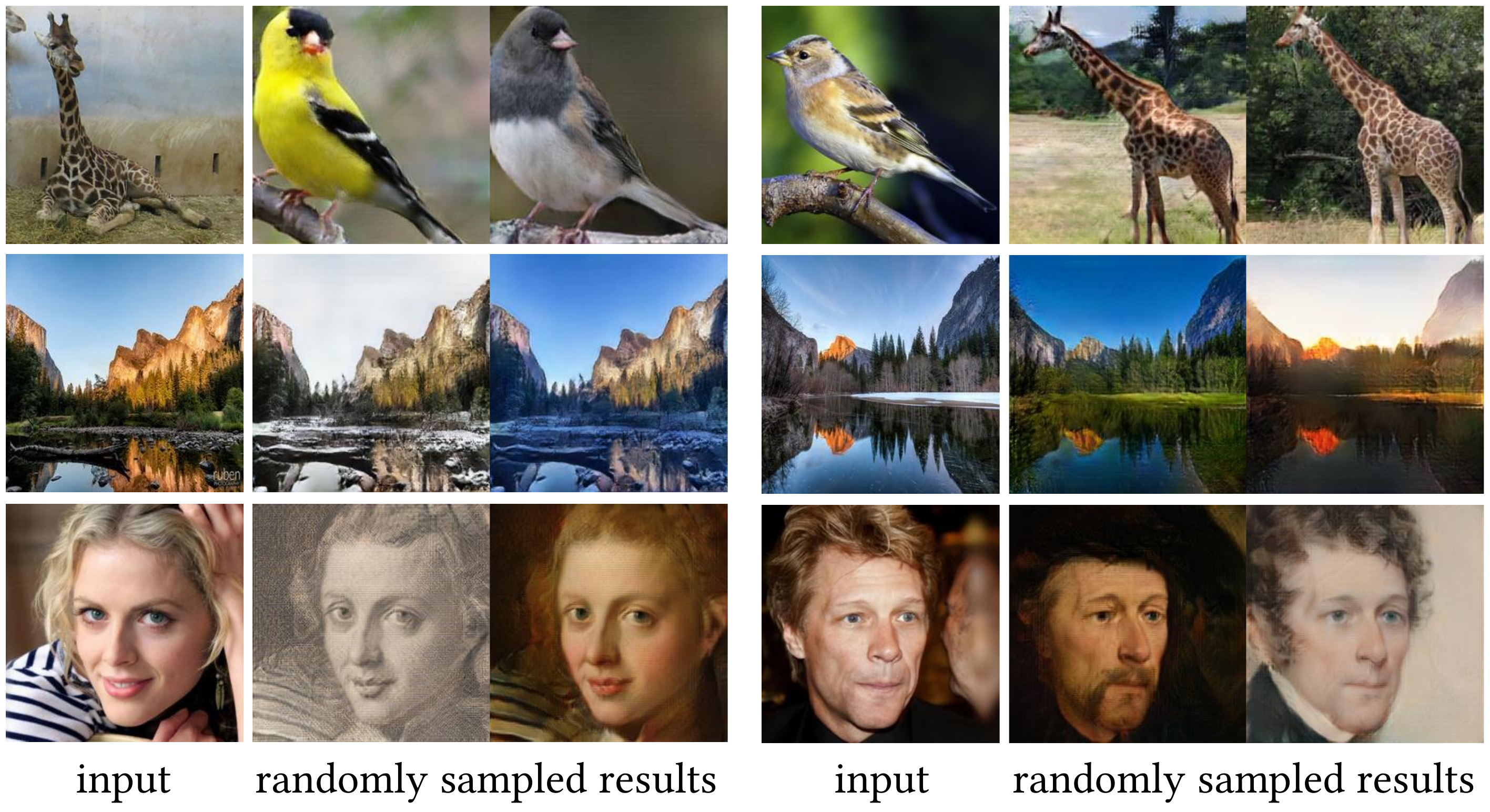}%\vspace{-2mm}
\caption{Applicability to domains beyond BigGAN: (\textit{top}) Giraffe $\leftrightarrow$ Bird,  (\textit{middle}) Summer $\leftrightarrow$ Winter, (\textit{bottom}) Face $\rightarrow$ Art.}%\vspace{-1mm}
\label{fig:generalization}
\end{figure}

\noindent
\textbf{Generalization to domains beyond BigGAN.}
Figure~\ref{fig:generalization} shows three applications of species translation, season transfer and facial stylization.~Even if MS-COCO giraffes~\cite{lin2014microsoft}, Yosemite landscapes~\cite{huang2018multimodal} and Art portraits~\cite{karras2020training} are not within the ImageNet 1,000 classes and are not observed by the content encoder in Stage I, our method can well support these domains and generate realistic results.

\begin{figure}[t]
\centering
    \includegraphics[width=1\linewidth]{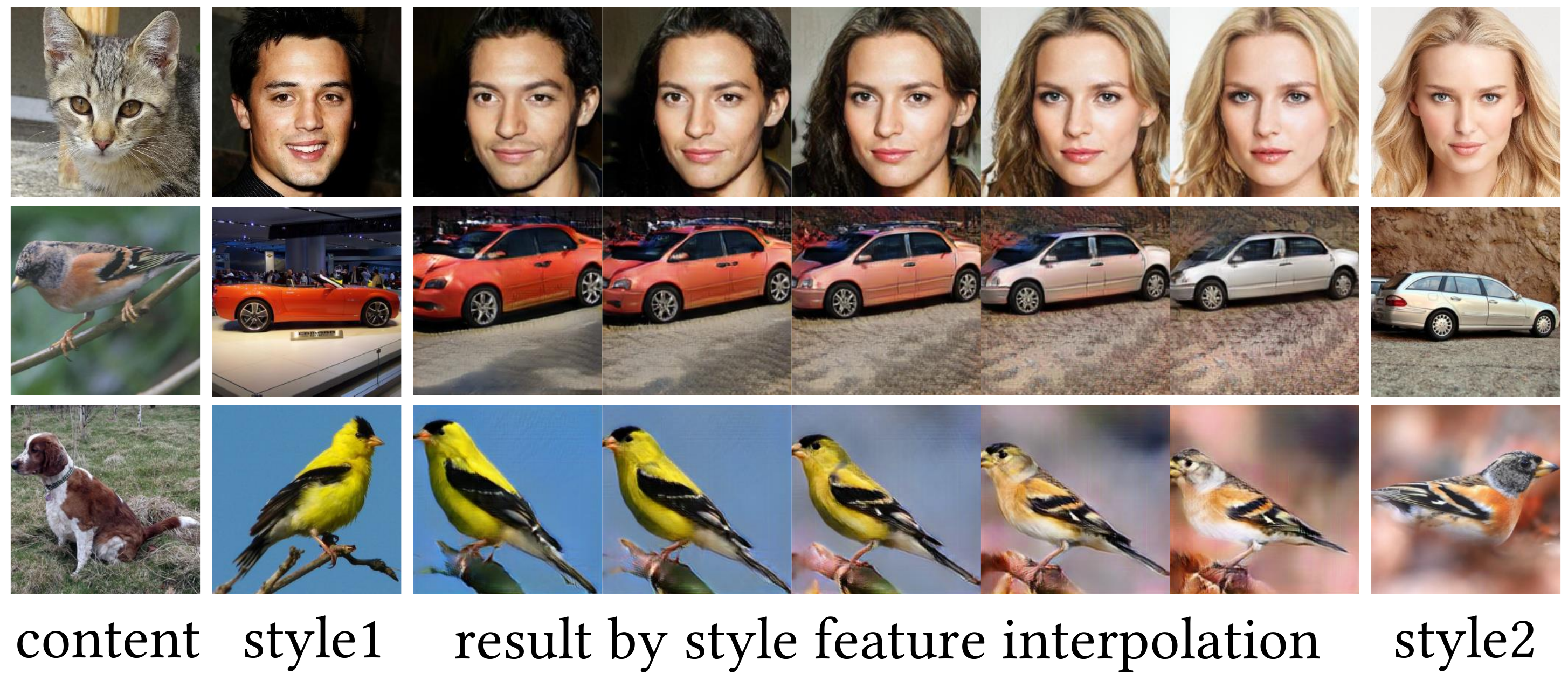}%\vspace{-2mm}
\caption{Style blending.}%\vspace{-1mm}
\label{fig:interpolation1}
\end{figure}

\noindent
\textbf{Style blending.} As shown in Fig.~\ref{fig:interpolation1}, we perform a linear interpolation to style feature, and observe smooth changes along with the latent space from one to another.

\begin{figure}[t]
\centering
    \includegraphics[width=1\linewidth]{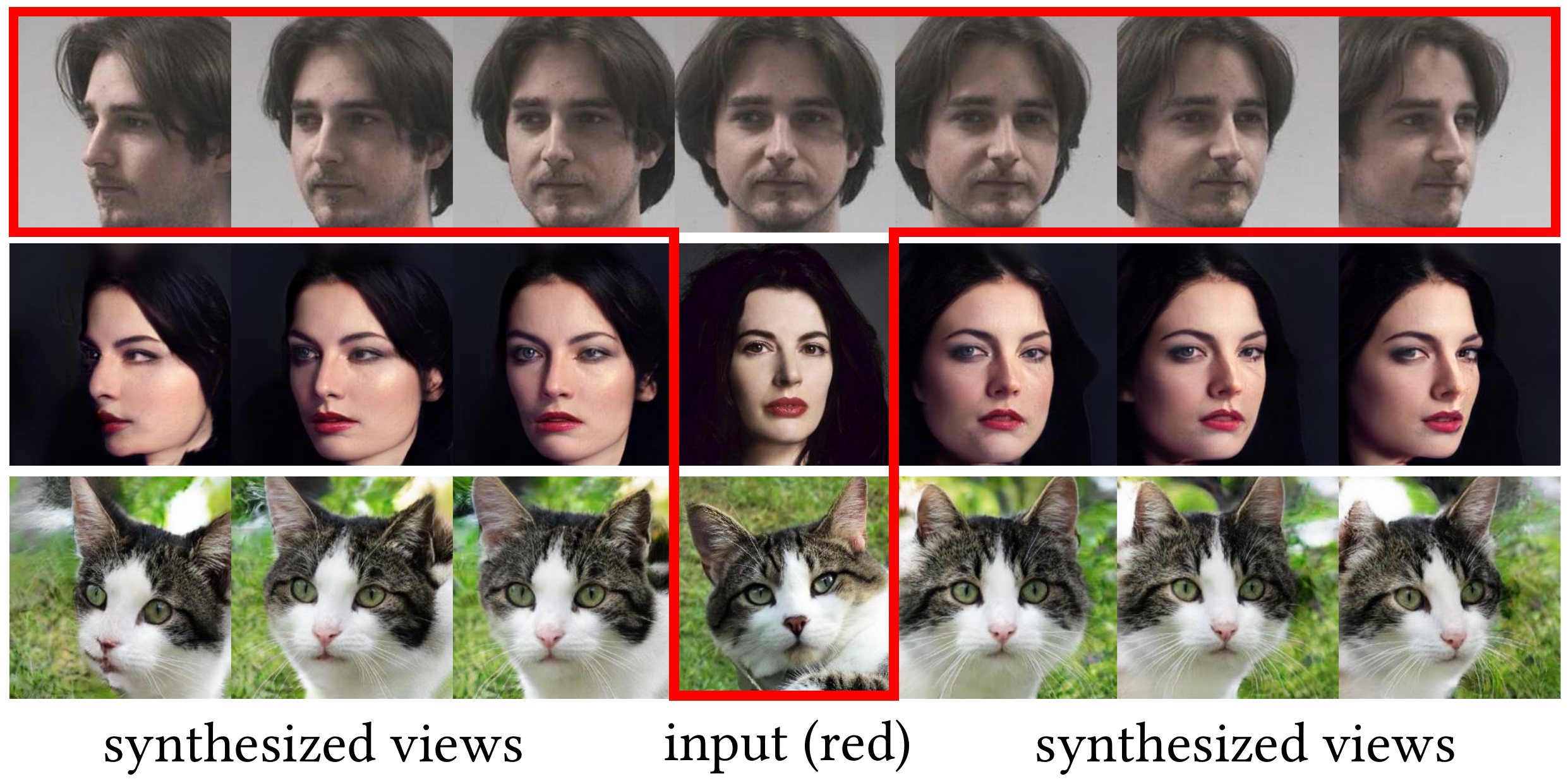}%\vspace{-2mm}
\caption{Unseen view synthesis.}%\vspace{-1mm}
\label{fig:interpolation2}
\end{figure}

\noindent
\textbf{Unseen view synthesis.}
Our exemplar-guided framework allows unseen view synthesis. Figure~\ref{fig:interpolation2} shows our synthesized realistic human and cat faces in various pan angles according to the reference faces from the Head Pose Image Database~\cite{gourier2004estimating}.

\subsection{Limitations}
\label{sec:limitation}

\begin{figure}[t]
\centering
\includegraphics[width=1\linewidth]{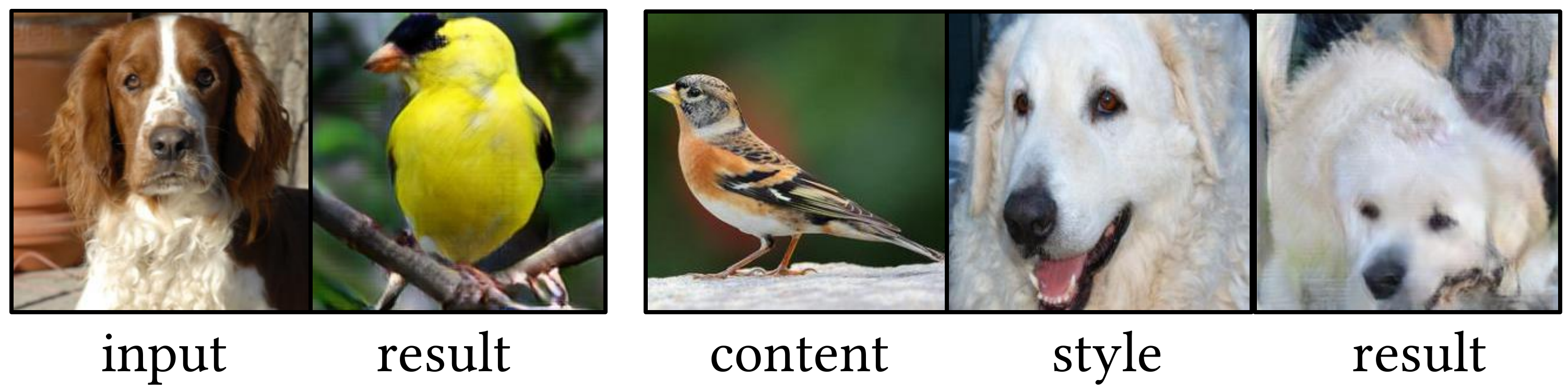}%\vspace{-2mm}
\caption{Failure cases caused by (\textit{left}) imbalanced training data, and (\textit{right}) scale mismatch of the content and style objects.}%\vspace{-1mm}
\label{fig:limitation}
\end{figure}

Figure~\ref{fig:limitation} gives two typical failure cases of our method.
First, our method fails to generate a bird sharing the same head direction as the dog, due to the lack of training images of birds looking directly at the camera.
Therefore, special attention should be taken when applying this method to applications where the possible data imbalance issue might lead to biased results towards minority groups  in the dataset.
Second, when the objects in the content and style images have very different scales, some appearance features cannot be rendered correctly.

\section{Conclusion}
\label{sec:conclusion}
\vspace{-1mm}

In this paper, we explore the use of GAN generative prior to build a versatile UNIT framework.
We show that the proposed two-stage framework is able to characterize content correspondences at a high semantic level for challenging multi-modal translations between distant domains.
An advantage is that such content correspondences can be discovered with only domain supervision (\ie, only knowing the domain each image belongs to).
We further find in Sec.~\ref{sec:ablation} that fine-level correspondences are learned merely via a generation task.
It poses a potential of Learning by Generation: building object relationships by generating and transforming them.
With well-designed dynamic skip connection modules and training objectives, our model can realize content correspondences at different granularities, thus enabling a trade-off between the content consistency and the style consistency.
Moreover, we show that such correspondences could be further improved by combining both generative priors and discriminative priors via semi-supervised learning.

\bibliographystyle{IEEEtran}
\bibliography{egbib}

\ifCLASSOPTIONcaptionsoff
  \newpage
\fi



%\bibliographystyle{IEEEtran}


%\iffalse

\begin{IEEEbiography}[{\includegraphics[width=1in,height=1.25in,clip,keepaspectratio]{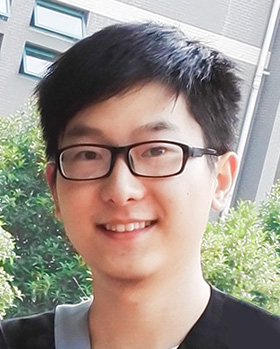}}]{Shuai Yang}(S'19-M'20) received the B.S. and Ph.D. degrees (Hons.) in computer science from Peking University, Beijing, China, in 2015 and 2020, respectively.
He is currently a Research Assistant Professor with the S-Lab, Nanyang Technological University, Singapore.
Dr. Yang was a postdoctoral research fellow at Nanyang Technological University, from Oct. 2020 to Feb. 2023.
He was a Visiting Scholar with the Texas A\&M University, from Sep. 2018 to Sep. 2019.
He was a Visiting Student with the National Institute of Informatics, Japan, from Mar. 2017 to Aug. 2017.
He received the IEEE ICME 2020 Best Paper Awards and IEEE MMSP 2015 Top10\% Paper Awards.
His current research interests include image stylization and image translation.
\end{IEEEbiography}

\begin{IEEEbiography}[{\includegraphics[width=1in,height=1.4in,clip,keepaspectratio]{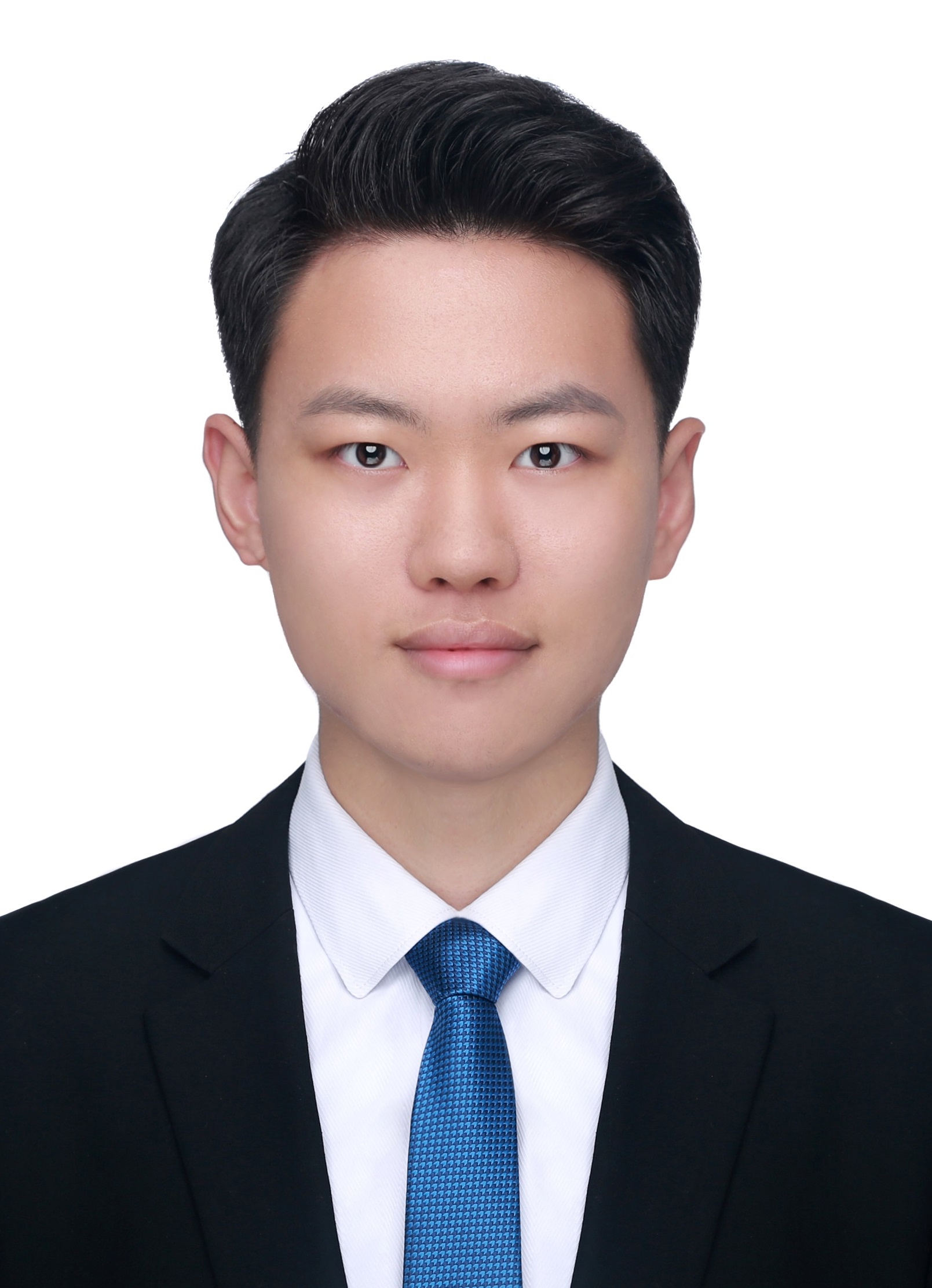}}]{Liming Jiang} is currently a Ph.D. candidate at S-Lab, Nanyang Technological University, Singapore.
His research focuses on computer vision, generative models, image and video synthesis.
Liming co-organized the DeeperForensics Challenge 2020, ECCV 2020 SenseHuman Workshop, and OpenMMLab Workshop 2021.
He was one of the main developers of MMEditing, a popular open-source image and video editing toolbox.
He obtained the ACM-ICPC Asia Regional Contest Gold Medal in 2017.
\end{IEEEbiography}

\begin{IEEEbiography}[{\includegraphics[width=1in,height=1.25in,clip,keepaspectratio]{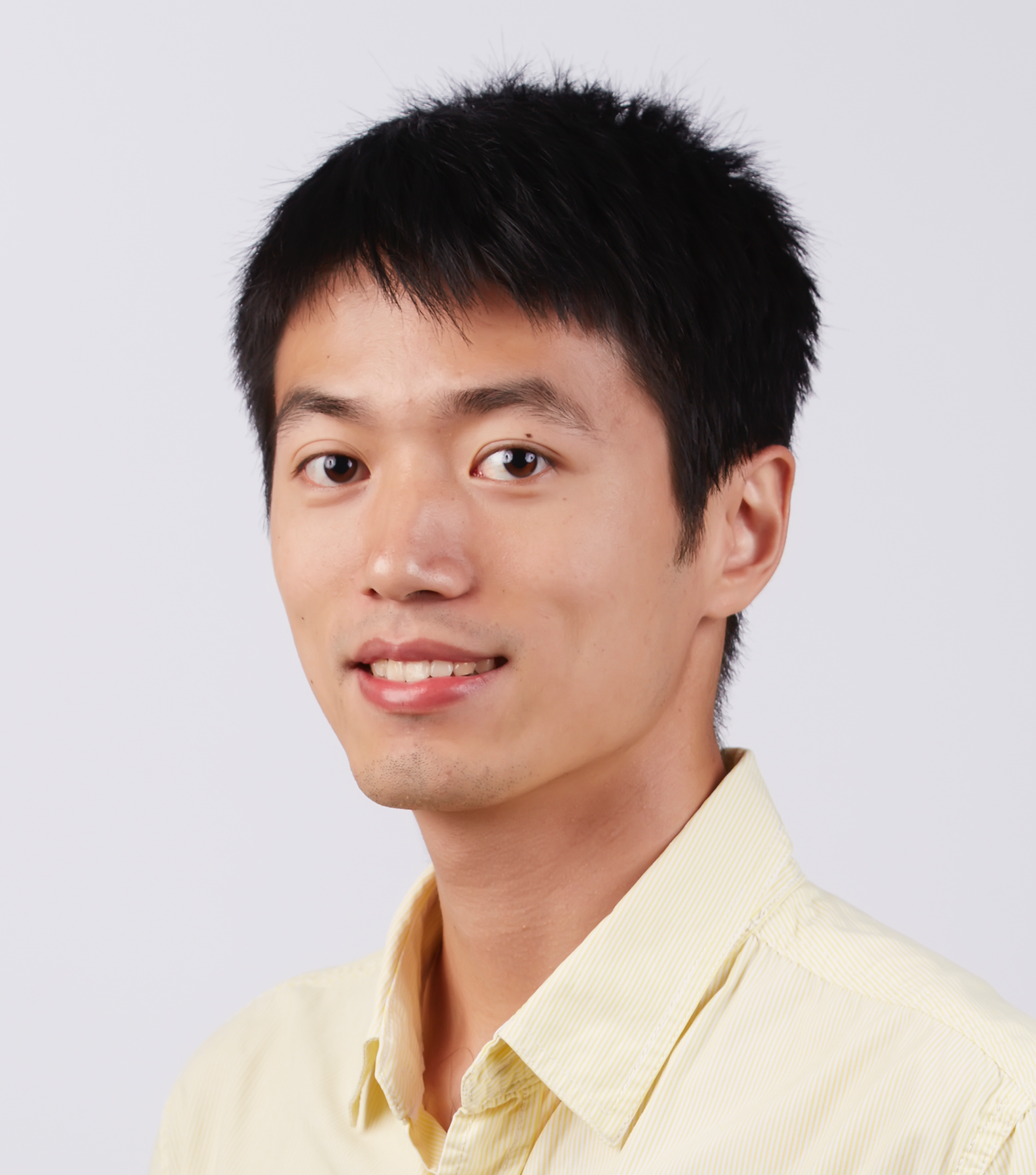}}]{Ziwei Liu} is currently a Nanyang Assistant Professor at Nanyang Technological University, Singapore. His research revolves around computer vision, machine learning and computer graphics. He has published extensively on top-tier conferences and journals in relevant fields, including CVPR, ICCV, ECCV, NeurIPS, ICLR, ICML, TPAMI, TOG and Nature - Machine Intelligence. He is the recipient of Microsoft Young Fellowship, Hong Kong PhD Fellowship, ICCV Young Researcher Award, HKSTP Best Paper Award and WAIC Yunfan Award. He serves as an Area Chair of CVPR, ICCV, NeurIPS and ICLR, as well as an Associate Editor of IJCV.
\end{IEEEbiography}

\begin{IEEEbiography}[{\includegraphics[width=1in,height=1.25in,clip,keepaspectratio]{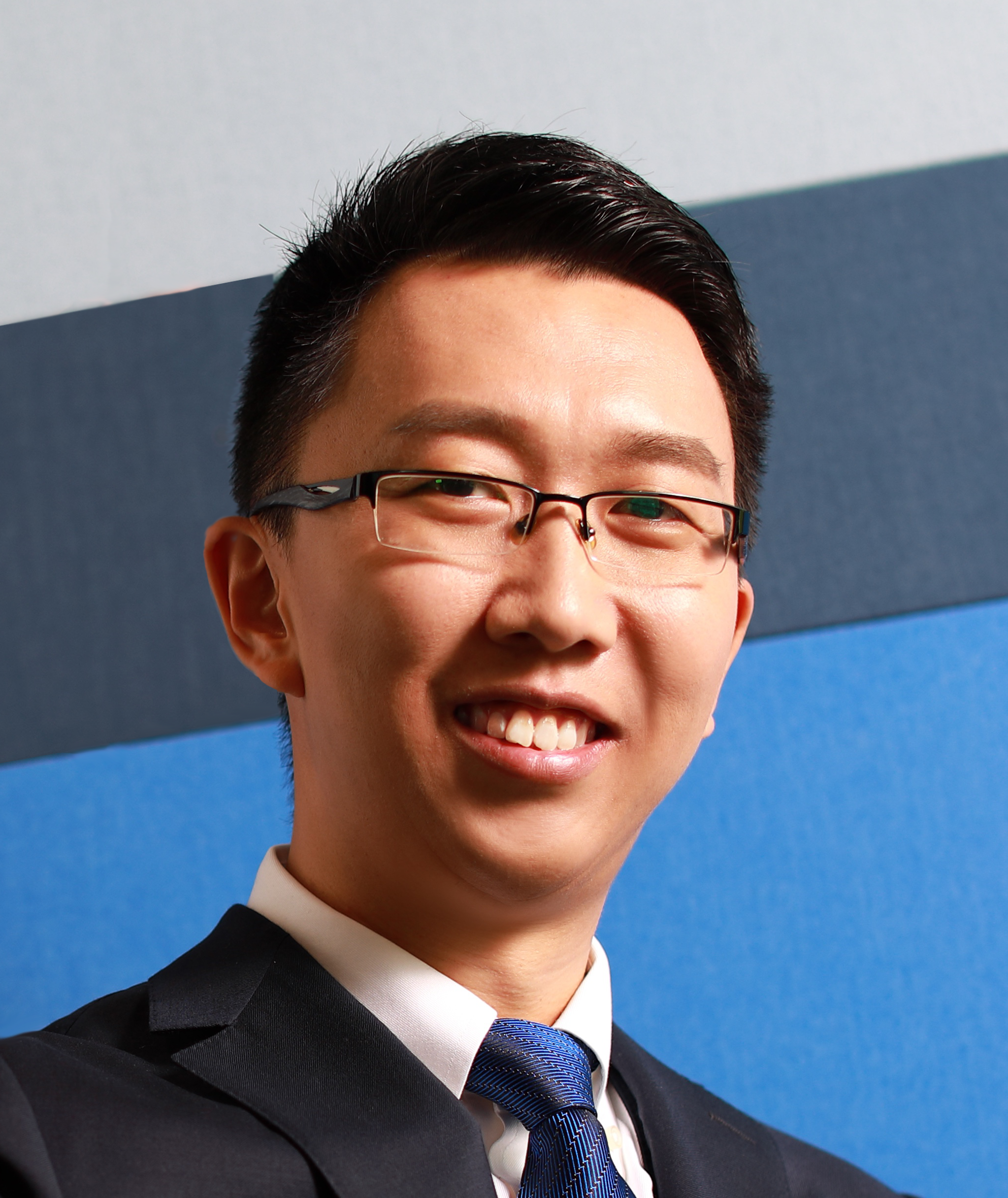}}]{Chen Change Loy} (Senior Member, IEEE) is currently a Nanyang Associate Professor with the School of Computer Science and Engineering, Nanyang Technological University, Singapore. He received the PhD degree in computer science from the Queen Mary University of London, in 2010. Prior to joining NTU, he served as a research assistant professor with the Department of Information Engineering, The Chinese University of Hong Kong, from 2013 to 2018. His research interests include computer vision and deep learning with a focus on image/video restoration and enhancement, generative tasks, and representation learning. He serves as an associate editor of the IEEE Transactions on Pattern Analysis and Machine Intelligence and the International Journal of Computer Vision. He also serves/served as an Area Chair of top conferences such as ICCV, CVPR and ECCV.
\end{IEEEbiography}

%\iffalse
%\fi
%\newpage


%\vfill

\end{document}